\documentclass{article}

\PassOptionsToPackage{numbers, sort&compress}{natbib}

\usepackage[preprint]{neurips_2024}

\usepackage[utf8]{inputenc} 
\usepackage[T1]{fontenc}    
\usepackage{hyperref}       
\usepackage{url}            
\usepackage{booktabs}       
\usepackage{amsfonts}       
\usepackage{nicefrac}       
\usepackage{microtype}      
\usepackage{xcolor}         
\usepackage{kotex}
\usepackage{lipsum}
\usepackage{pifont}
\usepackage[normalem]{ulem}
\useunder{\uline}{\ul}{}
\newcommand{\cmark}{\ding{51}}%
\newcommand{\xmark}{\ding{55}}%
\usepackage{mathtools}

\usepackage{array}
\newcolumntype{P}[1]{>{\centering\arraybackslash}p{#1}}
\newcolumntype{M}[1]{>{\centering\arraybackslash}m{#1}}

\usepackage{xspace}

\makeatletter
\DeclareRobustCommand\onedot{\futurelet\@let@token\@onedot}
\def\@onedot{\ifx\@let@token.\else.\null\fi\xspace}

\def\eg{\emph{e.g}\onedot} 
\def\ie{\emph{i.e}\onedot}

\makeatother

\usepackage{cleveref}
\usepackage{caption}

\usepackage{amsmath,amsfonts,bm}

\def\eqref#1{equation~\ref{#1}}

\def\1{\bm{1}}

\DeclareMathAlphabet{\mathsfit}{\encodingdefault}{\sfdefault}{m}{sl}
\SetMathAlphabet{\mathsfit}{bold}{\encodingdefault}{\sfdefault}{bx}{n}

\def\gD{{\mathcal{D}}}

\def\gL{{\mathcal{L}}}

\def\sR{{\mathbb{R}}}

\def\rmF{{\mathbf{F}}}

\newcommand{\Cov}{\mathrm{Cov}}

\title{Regularized Training with Generated Datasets for Name-Only Transfer of Vision-Language Models}

\author{
  Minho Park \quad Sunghyun Park \quad Jooyeol Yun \quad Jaegul Choo \\
  Kim Jaechul Graduate School of AI, KAIST \\
  \texttt{\{m.park, blizzard072, jchoo\}@kaist.ac.kr} \\
  \texttt{psh01087@gmail.com} \\
}

\begin{document}

\maketitle

\begin{abstract}
Recent advancements in text-to-image generation have inspired researchers to generate datasets tailored for perception models using generative models, which prove particularly valuable in scenarios where real-world data is limited.
In this study, our goal is to address the challenges in fine-tuning vision-language models (\eg, CLIP) on generated datasets.
Specifically, we aim to fine-tune vision-language models to a specific classification model without access to any real images, also known as name-only transfer.
However, despite the high fidelity of generated images, we observe a significant performance degradation when fine-tuning the model using the generated datasets due to the domain gap between real and generated images.
To overcome the domain gap, we provide two regularization methods for training and post-training, respectively.
First, as a post-training regularization, we leverage the domain-agnostic knowledge from the original pre-trained vision-language model via weight-space ensemble between the original model and the model fine-tuned on the generated dataset.
Secondly, we reveal that fine-tuned models with high feature diversity score high performance in the real domain, which indicates that increasing feature diversity prevents learning the generated domain-specific knowledge.
Thus, as a train-time regularization, we encourage feature diversity by providing additional regularization.
Extensive experiments on various classification datasets and various text-to-image generation models demonstrate that our analysis and regularization techniques effectively mitigate the domain gap, which has long been overlooked, and enable us to achieve state-of-the-art performance by training with generated images.
Code is available at \url{https://github.com/pmh9960/regft-for-gen}
\end{abstract}

\section{Introduction}
\label{sec:intro}

Recent advances in text-to-image generation have achieved remarkable success in producing high-quality images and capturing textual conditions~\cite{ldm, imagen, glide, dalle, sdxl}.
In response to these developments, researchers have begun exploring ways to enhance the performance in perception tasks (\eg, classification) by generating datasets from such powerful generative models~\cite{is_synthetic, cafo, susx, generative_robustness, sddm, imagenet_sd, stablerep}.
Specifically, image-label pairs for classification can be constructed using pre-trained text-to-image generation models by conditioning the models to generate ``A photo of a \texttt{[class name]}''.
The generated datasets are particularly valuable when real-world samples are insufficient.
In this study, we investigate the effective utilization of the generated datasets for fine-tuning vision-language models (\eg, CLIP) in the name-only transfer scenario that solely relies on class names and does not have access to real images to classify images.

\begin{figure}[t]
    \centering
    \includegraphics[width=0.82\columnwidth]{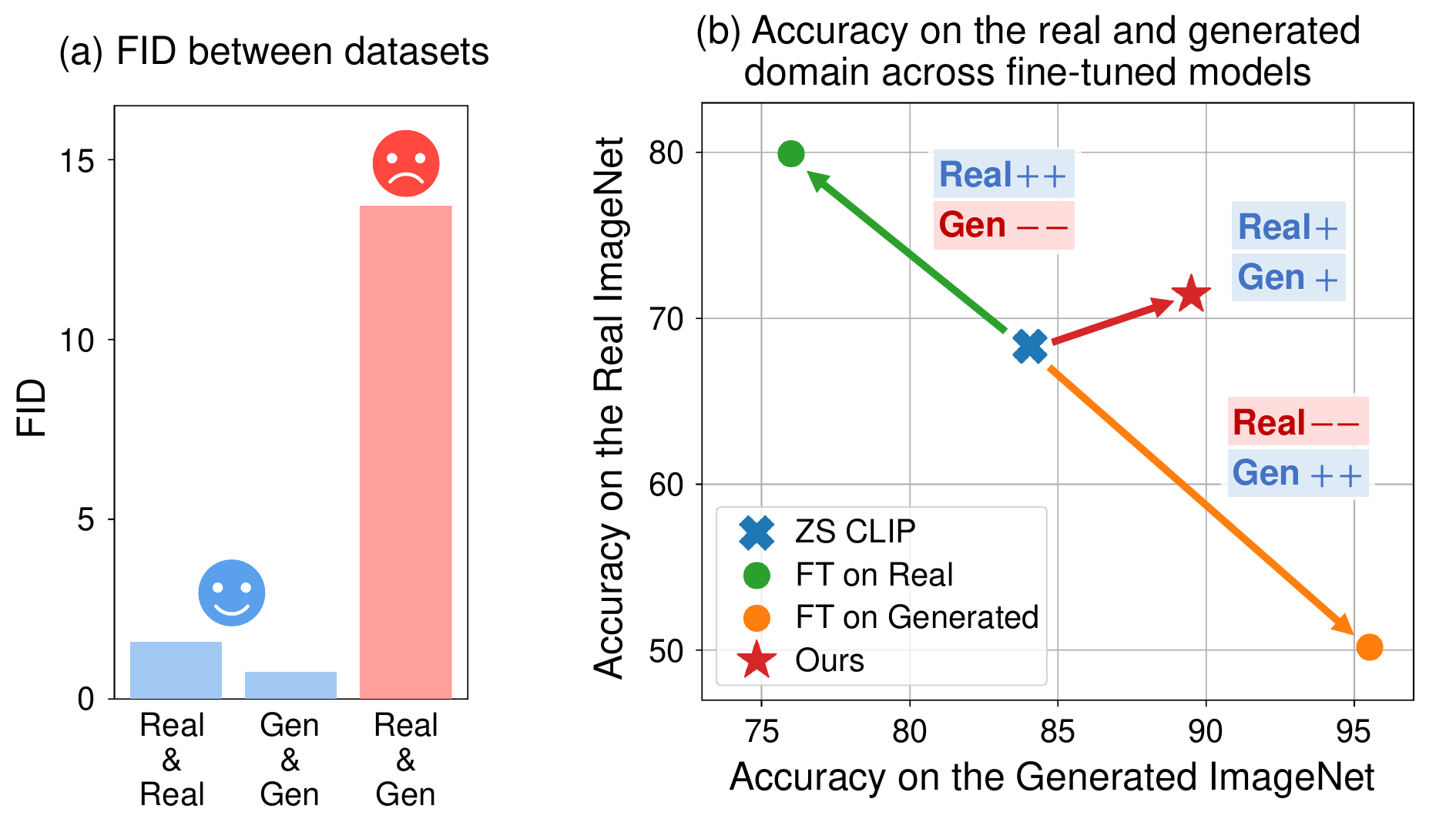}
    \caption{
    (a) Fr\'{e}chet Inception Distance (FID)~\cite{fid} of the intra-domain and inter-domain represents the significant domain gap between the real and generated images.
    (b) Accuracy of real and generated ImageNet~\cite{imagenet} across the original pre-trained vision-language model (\eg, CLIP~\cite{clip}) and the fine-tuned models.
    Fine-tuning on the specific domain often leads to performance degradation of the other domain, as shown in both real and generated domains.
    In this study, we aim to improve the real-domain accuracy by overcoming the domain gap with regularization techniques.
    }
    \vspace{-0.4cm}
    \label{fig:intro_fid}
\end{figure}


Despite the high fidelity of generated images, a notable domain gap persists between real and generated images.
We quantify this gap using the Fr\'{e}chet Inception Distance (FID)~\cite{fid}, which assesses the disparity between image sets. 
As illustrated in \Cref{fig:intro_fid} (a), the gap between inter-domain (\eg, Real-Gen) is significantly larger than the gap between intra-domain (\eg, Real-Real, Gen-Gen).
This domain gap leads to practical consequences when fine-tuning classifiers on the generated dataset.
As shown in \Cref{fig:intro_fid} (b), fine-tuning classifiers on the specific dataset (\eg, generated) can improve the in-domain accuracy (\eg, generated), but often leads to performance degradation in the other domain (\eg, real).
Furthermore, since we observe that the accuracy of the generated dataset also degrades when fine-tuning it on the real dataset, \emph{we interpret the situation as a subject of domain gap rather than the generated dataset's inferiority or mislabeling}.

In this study, we introduce two regularization methods that are necessary to learn the task-specific information (\eg, classification) from the generated dataset while overcoming the domain gap.
The regularization methods can be applied at training and post-training, respectively.
While previous name-only transfer approaches have refrained from fine-tuning the entire classifier and resorted to lightweight feature adapter~\cite{is_synthetic, cafo, calip, susx} or prompt engineering~\cite{cupl, menon2022visual}, \emph{we take the initiative to enhance the entire CLIP classifier including the image encoder} as illustrated in \Cref{fig:overview}.

First, to overcome the domain gap, we leverage the domain-agnostic knowledge from the original pre-trained vision-language model (\eg, CLIP~\cite{clip}) for regularizing fine-tuned models at the post-training time.
We highlight that the simple ensemble approach can effectively learn additional task-specific knowledge from the generated dataset if we leverage the domain-agnostic knowledge from the original model, which is the most desirable property in the dataset generation literature.

Secondly, we introduce a training-time regularization to prevent learning the generated domain-specific knowledge when fine-tuning the classifier on the generated datasets.
Based on a thorough analysis of fine-tuning with the generated dataset, we find that fine-tuned models with high feature diversity exhibit strong performance in the real domain.
This suggests that increasing feature diversity can help prevent the model from learning domain-specific knowledge confined to the generated data.
Therefore, we encourage feature diversity via training-time regularization loss.

Our extensive experiments showcase significant performance improvements across 11 name-only transfer datasets spanning various text-to-image generation models.
Moreover, we further examine the wide applicability of our enhanced image encoder in the few-shot classification setting by simply replacing the original image encoder with ours, achieving state-of-the-art performance.


\vspace{1cm}

\section{Related Work}
\label{sec:rel}

\paragraph{Name-only transfer of vision-language models}

In recent years, there has been a notable shift towards training vision foundation models by incorporating natural language supervision, as highlighted in several studies~\cite{clip, align, basic, coca}.
Among these, CLIP~\cite{clip} stands out for creating a joint embedding space for images and texts through contrastive learning, leveraging a vast dataset of 400 million image-text pairs.
The name-only transfer is also pioneered by CLIP, which classifies images solely based on class names to assess similarity between images and texts.
To achieve better performance than CLIP in the name-only transfer, enriching text input with large language models~\cite{gpt3} has been researched~\cite{cupl,menon2022visual}.

\vspace{-0.1cm}
\paragraph{Dataset generation for name-only transfer}

More recently, there has been significant progress in constructing a generated dataset by harnessing pre-trained text-to-image models in the name-only transfer of vision-language models~\cite{is_synthetic, cafo, susx}.
Specifically, these studies generated images from class names using textual input such as ``A photo of a \texttt{[class name]}'' to create the synthetic classification dataset.
However, we reveal that fine-tuning the entire classifier with the generated dataset degrades the performance on the real dataset due to the domain gap as depicted in \Cref{fig:intro_fid}.

\begin{figure*}[t]
    \centering
    \includegraphics[width=\linewidth]{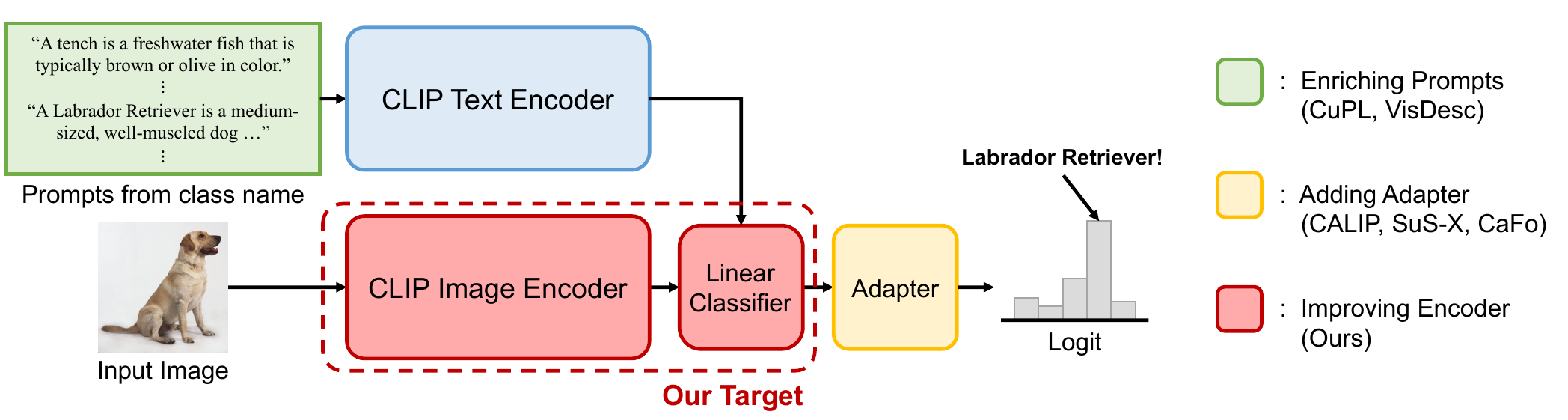}
    \vspace{-0.5cm}
    \caption{
    Overview of the architecture for name-only transfer of vision-language models (\eg, CLIP~\cite{clip}). 
    While the preceding approaches focused on {\color[RGB]{112, 173, 71} enriching prompts (green)} and {\color[RGB]{240, 180, 0} adapters (yellow)}, we aim to {\color[RGB]{192, 0, 0} fine-tuning the CLIP image encoder (red)} with generated datasets.
    }
    \label{fig:overview}
\end{figure*}

To prevent the overfitting to the generated dataset, previous approaches often bypassed fine-tuning the CLIP image encoder and employed techniques such as linear probing~\cite{is_synthetic} and adapters~\cite{cafo, susx}, which are commonly used in few-shot classification scenario~\cite{clip_adapter, tip_adapter}.
Although the CLIP image encoder holds notable potential for enhancing classification performance, it remains relatively under-explored.
In this paper, our target is the CLIP image encoder, which is the orthogonal research direction with the previous name-only transfer approaches as illustrated in \Cref{fig:overview}.

\vspace{-0.1cm}
\paragraph{Fine-tuning CLIP with limited real datasets}

Besides the name-only transfer, training CLIP with an insufficient dataset has also been researched in various ways.
CoOp~\cite{coop}, CLIP-Adapter~\cite{clip_adapter}, and Tip-Adapter~\cite{tip_adapter} enhance the CLIP framework by freezing the image encoder and introducing small trainable modules.
On the other hand, LPFT~\cite{lpft} introduces a two-step approach involving linear probing followed by full fine-tuning, while WiSE-FT~\cite{wiseft} proposes a weight-space ensemble by blending weights between the zero-shot model and the fine-tuned model.
Specifically, the weight-space ensemble has demonstrated significant efficacy in the natural distribution scenario (\eg, the photograph domain to the sketch domain).
Notwithstanding these advancements, the impact of fine-tuning CLIP on a generated dataset remains unexplored.
Thus, this study endeavors to examine the enhancements to CLIP through the incorporation of a generated dataset.

\vspace{-0.1cm}
\paragraph{Representation Learning by Diversifying Features}

In the recent past, self-supervised learning has achieved huge success within the invariance learning framework~\cite{simclr, vicreg, duality, caron2021emerging, zbontar2021barlow, oquab2023dinov2, he2020momentum, grill2020bootstrap, caron2020unsupervised, chen2021exploring}.
Several studies emphasize the importance of increasing feature diversity in self-supervised learning and transfer learning, resulting in significant performance improvements in downstream tasks~\cite{vicreg, vcr, zbontar2021barlow}.
These approaches specifically aim to prevent informational collapse by reducing the off-diagonal elements of the covariance matrix over a batch while diversifying each element by enlarging the diagonal elements of the covariance matrix.
Inspired by these pre-training approaches, we introduce two orthogonal metrics that assess the classifier's change when fine-tuning on the generated datasets.

\section{Proposed Method}
\label{sec:method}

\begin{figure*}[t]
    \centering
    \includegraphics[width=\textwidth]{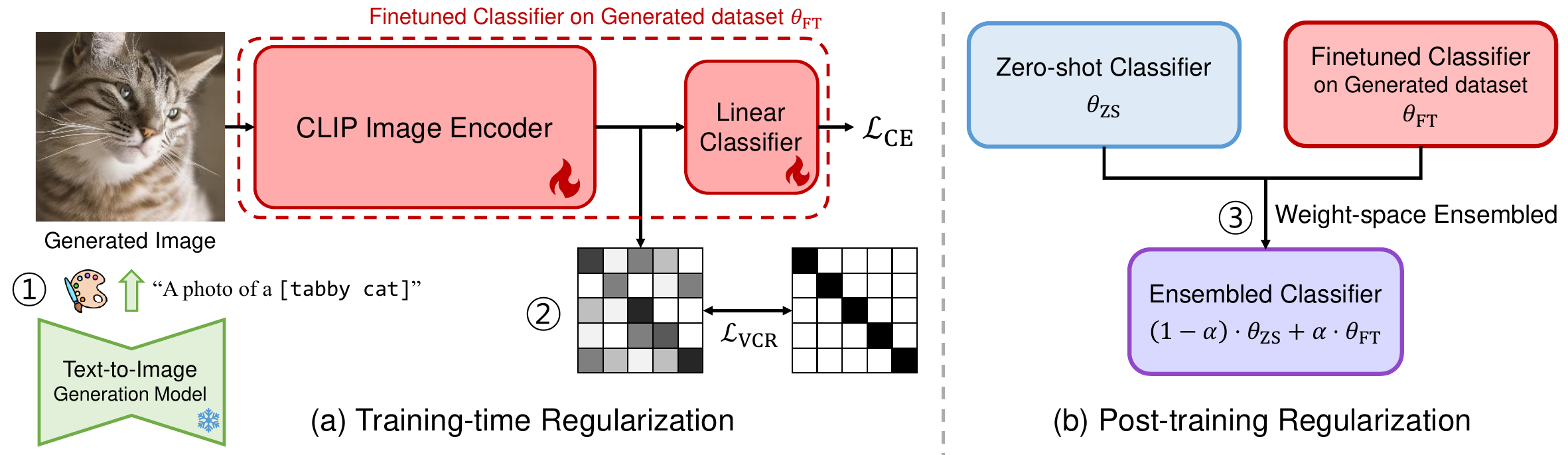}
    \caption{
    Overview of the proposed method.
    Initially, generated datasets are synthesized from textural conditions via text-to-image generation models.
    Subsequently, the entire classifier is fine-tuned on the generated dataset, employing cross-entropy loss ($\gL_\text{CE}$) with variance-covariance regularization ($\gL_\text{VCR}$).
    Lastly, a weight-space ensemble is performed to integrate the zero-shot classifier and the fine-tuned classifier.
    }
    \label{fig:method}
\end{figure*}

In this section, we provide our regularization methods based on our insight and analysis when fine-tuning the pre-trained vision-language model (\eg, CLIP) on the generated datasets.
The regularization methods are split into two parts: training-time and post-training regularization, as depicted in \Cref{fig:method}.

\subsection{Post-training Regularization: Weight-space Ensemble}
\label{sec:method_ensemble}

As illustrated in \Cref{fig:intro_fid} (b), the real-domain accuracy degrades when fine-tuning on the generated dataset without any regularization.
Importantly, we also emphasize that \emph{the accuracy of the generated dataset also decreases when fine-tuning on the real dataset}, indicating that the performance degradation may be due to a domain gap rather than the inferiority or mislabeling of the generated dataset.
Thus, we regularize the fine-tuned model with the generated dataset by leveraging the domain-agnostic property of the pre-trained vision-language model (\eg, CLIP~\cite{clip}).

Based on the interpretation, we conduct a weight-space ensemble~\cite{wiseft} of the fine-tuned classifier with the zero-shot CLIP classifier as a simple yet effective post-training regularization to leverage the domain-agnostic property of the zero-shot CLIP, as depicted in \Cref{fig:method} (b).
The weight-space ensemble is a simple linear interpolation between every parameter between two models ($\theta_{\text{ZS}}, \theta_\text{FT}$).
\begin{equation}
    \vspace{0.1cm}
    \text{WSE}(\theta_{\text{ZS}}, \theta_{\text{FT}}) = (1 - \alpha) \cdot \theta_{\text{ZS}} + \alpha \cdot \theta_{\text{FT}},
    \vspace{0.1cm}
\end{equation}
where $\alpha$ is a weight mixing coefficient determining the ensemble ratio between the classifiers.
The rationale behind this strategy is to leverage both the domain-agnostic property of the zero-shot CLIP classifier and the task-specific knowledge (\eg, classification) of the fine-tuned classifier from the generated dataset.
Although weight-space ensemble has originally been proposed to promote robustness in the natural distribution shift scenario, applying the technique to the models trained on generated datasets is under-explored. 
We bring attention to this technique by identifying our main obstacle in training with generated images as the domain gap, \emph{a perspective that has been overlooked}. 

By utilizing the domain-agnostic property of the zero-shot CLIP classifier, we can achieve task-specific knowledge from the generated datasets veiled by the domain gap between the real and generated images.
Furthermore, due to its simplicity and flexibility, it can be applied in other perception applications (\eg, detection, segmentation), which may suffer from the domain gap between the real and generated images to utilize the generated datasets.

\subsection{How Does Fine-tuning on Generated Dataset Alter Classifiers?}

In the following sections, we explore the correlation between feature diversity and performance experimentally, fine-tuning classifiers with various hyper-parameters on the generated dataset.
Specifically, we examine the feature diversity of the classifiers using two orthogonal metrics: magnitude diversity~($\gD_\text{Mag}$) and direction diversity~($\gD_\text{Dir}$).

\paragraph{Magnitude Diversity {\normalfont ($\gD_\text{Mag}$)}}

Magnitude diversity pertains to the range of values for each feature element.
We define the magnitude diversity of the classifier by utilizing a covariance matrix of the encoded features across dimensions.
Particularly, the magnitude diversity is defined as the largest eigenvalue $\lambda_{\text{max}}$ of the covariance matrix, which represents the magnitude of the principal component.
The following equations outline the calculation of the classifier's magnitude diversity by extracting $N$ test images into $D$-dimensional features (\ie, $\rmF \in \sR^{N \times D}$):
\begin{align}
    \Cov(\rmF) &= (\rmF - \mu_{\rmF})^T (\rmF - \mu_{\rmF}) \in \sR^{D\times D}, \\
    \label{eq:d_mag}
    \gD_\text{Mag} &\coloneqq \lambda_{\text{max}} ( \Cov(\rmF) ),
\end{align}
where $\mu_{\rmF} \in \sR^{D}$ denotes the mean vector of the rows of $\rmF$.

\begin{figure}[t]
    \centering
    \includegraphics[width=0.82\columnwidth]{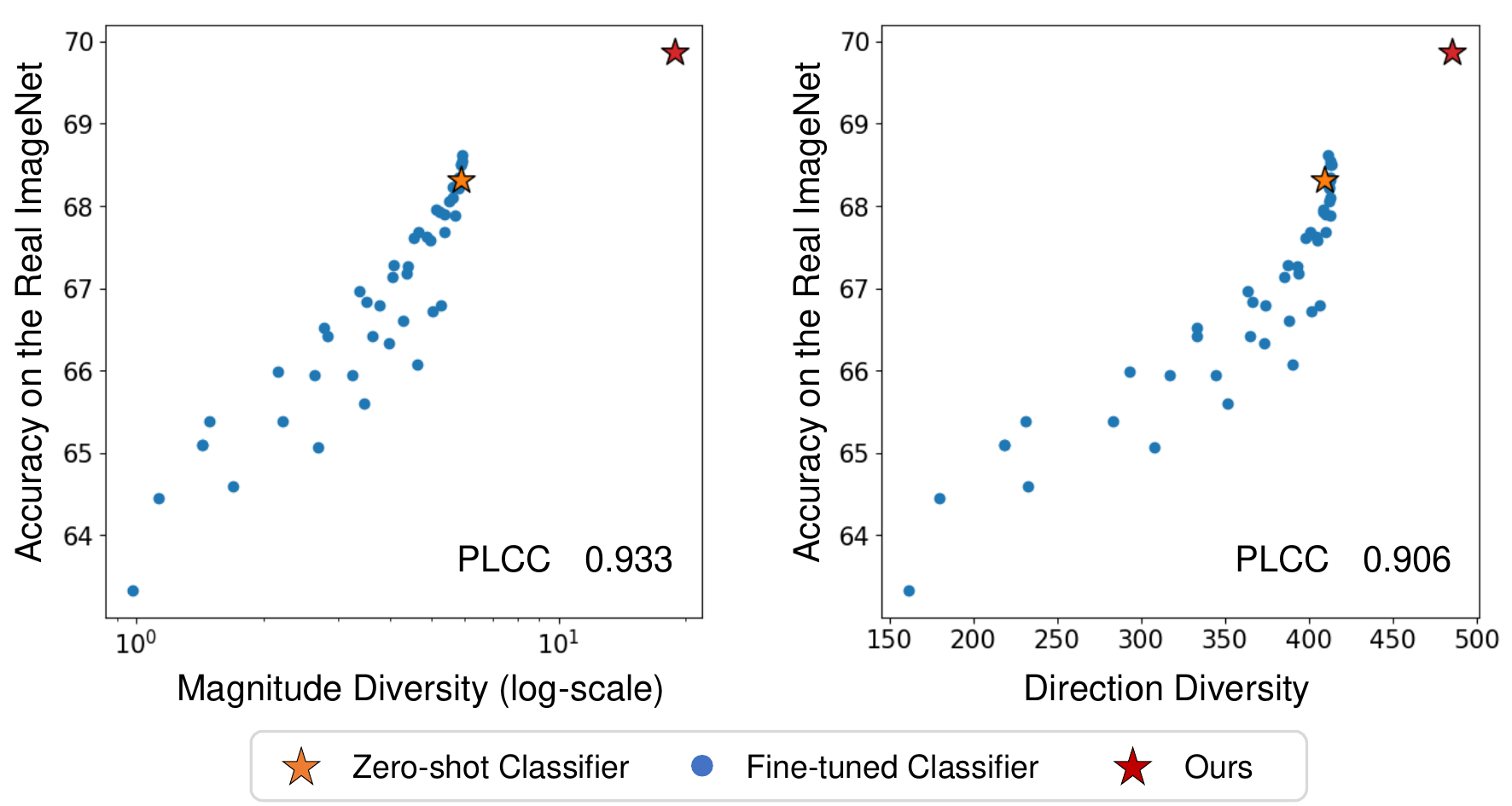}
    \caption{
    Evaluating magnitude diversity, direction diversity, and the real ImageNet~\cite{imagenet} accuracy of fine-tuned classifiers with the generated dataset.
    The results indicate a strong correlation between the diversity and robustness of the real domain.
    According to the observation, we successfully improved the performance in the real domain via both regularization methods.
    }
    \label{fig:intro_motivation}
    \vspace{-0.4cm}
\end{figure}

\paragraph{Direction Diversity {\normalfont ($\gD_\text{Dir}$)}}

Direction diversity concerns the variety of elements within the feature.
Since it also expresses how orthogonal the features are, we leverage the covariance matrix again.
We normalize the features with the square root of magnitude diversity~($\sqrt{\gD_\text{Mag}}$) before calculating the covariance matrix to discard the influence of magnitude diversity (\ie, set the magnitude diversity 1).
Direction diversity is defined by taking the inverse of the L2 norm of the off-diagonal elements.

The rationale behind the definition of direction diversity is that a small off-diagonal element of the covariance matrix indicates a low similarity between two feature dimensions, and low similarities across all dimensions indicate high diversity.
The following equations outline the calculation of the direction diversity from the same features:
\begin{align}
    \tilde{\rmF} &\coloneqq \frac{1}{\sqrt{\gD_\text{Mag}}} \rmF, \\
    \Cov(\tilde{\rmF}) &= (\tilde{\rmF} - \mu_{\tilde{\rmF}})^T (\tilde{\rmF} - \mu_{\tilde{\rmF}}) \in \sR^{D\times D}, \\
    \label{eq:d_dir}
    \gD_\text{Dir} &\coloneqq \frac{1}{\sqrt{\frac{1}{D (D-1)}\mathop{\sum \sum}_{i\neq j} (\Cov(\tilde{\rmF})_{ij})^2}},
\end{align}
where $\mu_{\tilde{\rmF}} \in \sR^{D}$ denotes the mean vector of the rows of $\tilde{\rmF}$.

\paragraph{Analysis}
We assess the defined diversity of various fine-tuned models with the generated dataset.\footnote{We generate 64 images for 1000 ImageNet~\cite{imagenet} classes by Stable Diffusion v2.1~\cite{ldm} with ``a photo of a \texttt{[class name]}'' for generated training-set and generate additional 5000 images for extracting features with the fine-tuned classifiers. The hyper-parameter set is constructed by varying learning rate $\in \{3\times 10^{-6}, 1\times 10^{-7}, 3\times 10^{-7}\}$, batch size $\in \{32, 64, 128\}$, and epochs $\in \{1, 2, 3, 5, 10\}$.}
The correlation between both diversity and the real accuracy is demonstrated in \Cref{fig:intro_motivation}, and it is also quantified by the Pearson linear correlation coefficient (PLCC)~\cite{freedman2007statistics}.
\emph{We observed that the increasing performance on the real domain is highly related to the increasing feature diversity.}
Thus, we introduce an additional regularization to increase the feature diversity of classifiers during fine-tuning classifiers on the generated datasets.

\subsection{Training-time Regularization: Variance-Covariance Regularization}
\label{sec:method_vcr}

As illustrated in \Cref{fig:intro_motivation}, since classifiers that have high feature diversity often provide high real-domain accuracy, utilizing diversity metrics in training-time regularization is the straightforward approach to improve real-domain accuracy.
However, both diversity metrics lack a closed-form derivative since the largest eigenvalue of the covariance matrix is computed by solving characteristic polynomials~\cite{kato2013perturbation}.
Thus, we have to discard the eigenvalue term from $\gD_\text{Mag}$ and $\gD_\text{Dir}$ while maintaining the main property of the metrics, resulting in the regularization techniques for the self-supervised learning, known as variance-covariance regularization~\cite{vicreg, vcr}.
\begin{equation}
\label{eq:vcr}
    \gL_{\text{VCR}} = \lambda_\text{Var} \cdot \frac{1}{D} \sum_{i=1}^D \max (0, 1 - \sqrt{C_{ii}}) 
    + \lambda_\text{Cov} \cdot \frac{1}{D (D-1)} \mathop{\sum \sum}_{i\neq j} C_{ij}^2,
\end{equation}
where $C$ is the covariance matrix over a mini-batch, $D$ is the dimension of the embedding features, and $\lambda_\text{Var}, \lambda_\text{Cov}$ are the strength of the variance and covariance regularization, respectively.

\paragraph{Interpreting VCR in terms of {\normalfont $\gD_\text{Mag}, \gD_\text{Dir}$}}

The first term of the variance-covariance regularization increases the sum of the diagonal elements of the covariance matrix, which is equal to the sum of the eigenvalues~\cite{strang2022introduction}.
Since increasing the largest eigenvalue can be circumvented by increasing the sum of all eigenvalues, increasing the magnitude diversity can be aimed by the first term.
Furthermore, the second term reduces the off-diagonal elements of the covariance matrix, which can effectively increase the direction diversity.
As a result, the variance-covariance regularization successfully increases both diversity and improves the real-domain performance as shown in \Cref{fig:intro_motivation}.


\section{Experiments}
\label{sec:exp}

\subsection{Implementation Details}

Our experiments are conducted across three text-to-image generation models: DALL-E~\cite{dalle, dalle_mini}, Stable Diffusion 2.1~\cite{ldm}, and Stable Diffusion XL~\cite{sdxl}.
We generate 64 images per class, with a 5.0 guidance scale for achieving high-fidelity images.
The input text prompts for text-to-image generation models for each dataset are demonstrated in \Cref{sec:supp_hand_written_prompts}, which are given by previous name-only transfer research~\cite{cafo, is_synthetic, susx}.
We employ CLIP ViT-B/16~\cite{clip, vit} as a pre-trained classifier.
Additional implementation details for fine-tuning classifiers and the experiments for different visual backbones of CLIP, including ResNet-50~\cite{resnet} were reported in \Cref{sec:supp_various_clip_results}.

\subsection{Name-only Transfer of Vision-Language Models}
\label{sec:exp_zeroshot}

\paragraph{Datasets}
We carried out name-only transfer experiments across 11 datasets, covering a diverse range of objects, scenes, and fine-grained categories.
These datasets include ImageNet~\cite{imagenet}, Caltech101~\cite{caltech101}, DTD~\cite{dtd}, EuroSAT~\cite{eurosat}, FGVCAircraft~\cite{fgvc_aircraft}, Flowers102~\cite{flowers102}, Food101~\cite{food101}, OxfordPets~\cite{oxford_pets}, StanfordCars~\cite{stanford_cars}, SUN397~\cite{sun397}, and UCF101~\cite{ucf101}.

\paragraph{Experimental settings and baselines}

We compared our method with the following name-only transfer baselines.
To begin with, we assess the zero-shot CLIP~\cite{clip} with various textural templates, following the ensemble approach for text embeddings.
CALIP~\cite{calip} incorporates spatial information of the CLIP image features via parameter-free attention.
CuPL~\cite{cupl}, leveraging large-language models~\cite{gpt3} to enrich prompts for CLIP text encoder, provides GPT prompts, so we evaluate classification performance using these given prompts.
CaFo~\cite{cafo} initializes with the GPT prompts~\cite{gpt3, cupl} and fine-tunes the adapter using generated datasets.
SuS-X~\cite{susx}, utilizing generated images as support set without additional fine-tuning, for a fair comparison, we use the provided generated datasets from DALL-E~\cite{dalle} by CaFo~\cite{cafo}.
Since our proposed method can easily integrate with other name-only transfer approaches by replacing the CLIP image encoder, we adopt GPT prompts~\cite{gpt3, cupl} for the initial classifier and integrate with the adapter method~\cite{cafo}.

\begin{table}[t!]
\centering
\caption{
Comparison of accuracy on 11 name-only transfer benchmarks.
Average indicates the average accuracy across the 11 datasets.
\underline{Underline} represents the highest accuracy achieved when utilizing generated datasets by DALL-E~\cite{dalle} given by CaFo~\cite{cafo}, while \textbf{Bold} indicates the highest accuracy achieved without any constraints.
}
\resizebox{\textwidth}{!}{
\begin{tabular}{@{}M{2.5cm}|M{2.2cm}M{2.2cm}M{2.2cm}M{2.2cm}M{2.2cm}M{2.2cm}@{}}
\toprule
\multicolumn{1}{l|}{} & \multicolumn{1}{c|}{Average} & ImageNet~\cite{imagenet} & Caltech101~\cite{caltech101} & DTD~\cite{dtd} & EuroSAT~\cite{eurosat} & FGVC~\cite{fgvc_aircraft} \\ \midrule
ZS CLIP & \multicolumn{1}{c|}{64.66} & 68.32 & 93.06 & 45.04 & 47.72 & 23.67 \\
CALIP & \multicolumn{1}{c|}{64.92} & 68.68 & 93.91 & 44.98 & 47.51 & 23.43 \\
CuPL & \multicolumn{1}{c|}{67.12} & 69.65 & 94.28 & 54.55 & 40.70 & 27.90 \\
SuS-X (DALL-E) & \multicolumn{1}{c|}{68.14} & 70.19 & 93.43 & 54.49 & \underline{48.94} & 27.81 \\
CaFo (DALL-E) & \multicolumn{1}{c|}{68.22} & 70.63 & \underline{94.85} & 54.96 & 48.56 & 28.47 \\
Ours (DALL-E) & \multicolumn{1}{c|}{\underline{69.31}} & \underline{70.87} & 94.40 & \underline{56.68} & 47.16 & \underline{\textbf{30.27}} \\
\midrule
CaFo (SDv2.1) & \multicolumn{1}{c|}{67.46} & 70.20 & 94.52 & 54.73 & 39.36 & 28.02 \\
CaFo (SDXL) & \multicolumn{1}{c|}{67.47} & 69.65 & 94.28 & 54.79 & 40.96 & 28.02 \\
Ours (SDv2.1) & \multicolumn{1}{c|}{70.26} & \textbf{71.43} & 94.93 & \textbf{58.10} & 56.21 & 28.50 \\
Ours (SDXL) & \multicolumn{1}{c|}{\textbf{71.04}} & 71.11 & \textbf{95.13} & 56.74 & \textbf{58.68} & 29.40 \\ \midrule
\multicolumn{1}{l|}{} & Flowers102~\cite{flowers102} & Food101~\cite{food101} & OxfordPets~\cite{oxford_pets} & StanfordCars~\cite{stanford_cars} & SUN397~\cite{sun397} & UCF101~\cite{ucf101} \\ \midrule
ZS CLIP & 66.10 & 83.85 & 87.11 & 66.29 & 65.13 & 65.00 \\
CALIP & 67.19 & 84.51 & 86.97 & 66.01 & 65.70 & 65.24 \\
CuPL & 69.47 & 86.35 & 90.54 & 66.62 & 68.00 & 70.29 \\
SuS-X (DALL-E) & 71.90 & 86.51 & 90.65 & 67.03 & 68.25 & 70.37 \\
CaFo (DALL-E) & 69.83 & 86.38 & 90.43 & 67.31 & 68.52 & 70.47 \\
Ours (DALL-E) & \underline{\textbf{73.28}} & \underline{\textbf{86.53}} & \underline{91.91} & \underline{70.55} & \underline{69.40} & \underline{71.32} \\
\midrule
CaFo (SDv2.1) & 69.75 & 86.43 & 91.22 & 69.12 & 68.34 & 70.37 \\
CaFo (SDXL) & 69.79 & 86.35 & 90.87 & 69.21 & 68.00 & 70.29 \\
Ours (SDv2.1) & 71.30 & 86.41 & 92.40 & 72.74 & \textbf{69.88} & 70.98 \\
Ours (SDXL) & 72.76 & 86.35 & \textbf{92.50} & \textbf{78.40} & 68.84 & \textbf{71.50} \\ \bottomrule
\end{tabular}
}
\label{tab:exp_zeroshot_accuracy}
\end{table}

\paragraph{Main results}
Our extensive experiments demonstrate that our method significantly and generally outperforms the name-only transfer baselines in 11 datasets, as illustrated in \Cref{tab:exp_zeroshot_accuracy}.
Our approach exhibited significant and general improvements on the DALL-E dataset, outperforming the zero-shot CLIP by +4.65 and the second-best approach, CaFo, by +1.11.
These results indicate that we successfully regularize the fine-tuning CLIP classifier when using the generated datasets.
Furthermore, as we extend the text-to-image generation model to SDv2.1 and SDXL, our method consistently outperforms across all datasets, whereas CaFo fails to show performance improvements with the change in generation models.
More discussion of the effects of changing the text-to-image generation models is available in \Cref{sec:supp_change_text_to_image_generator}.
The compatibility of our method with the other name-only transfer approaches is discussed in \Cref{sec:supp_compatibility}.

\subsection{Few-shot Classification}

\begin{figure}[t]
    \centering
    \includegraphics[width=0.8\columnwidth]{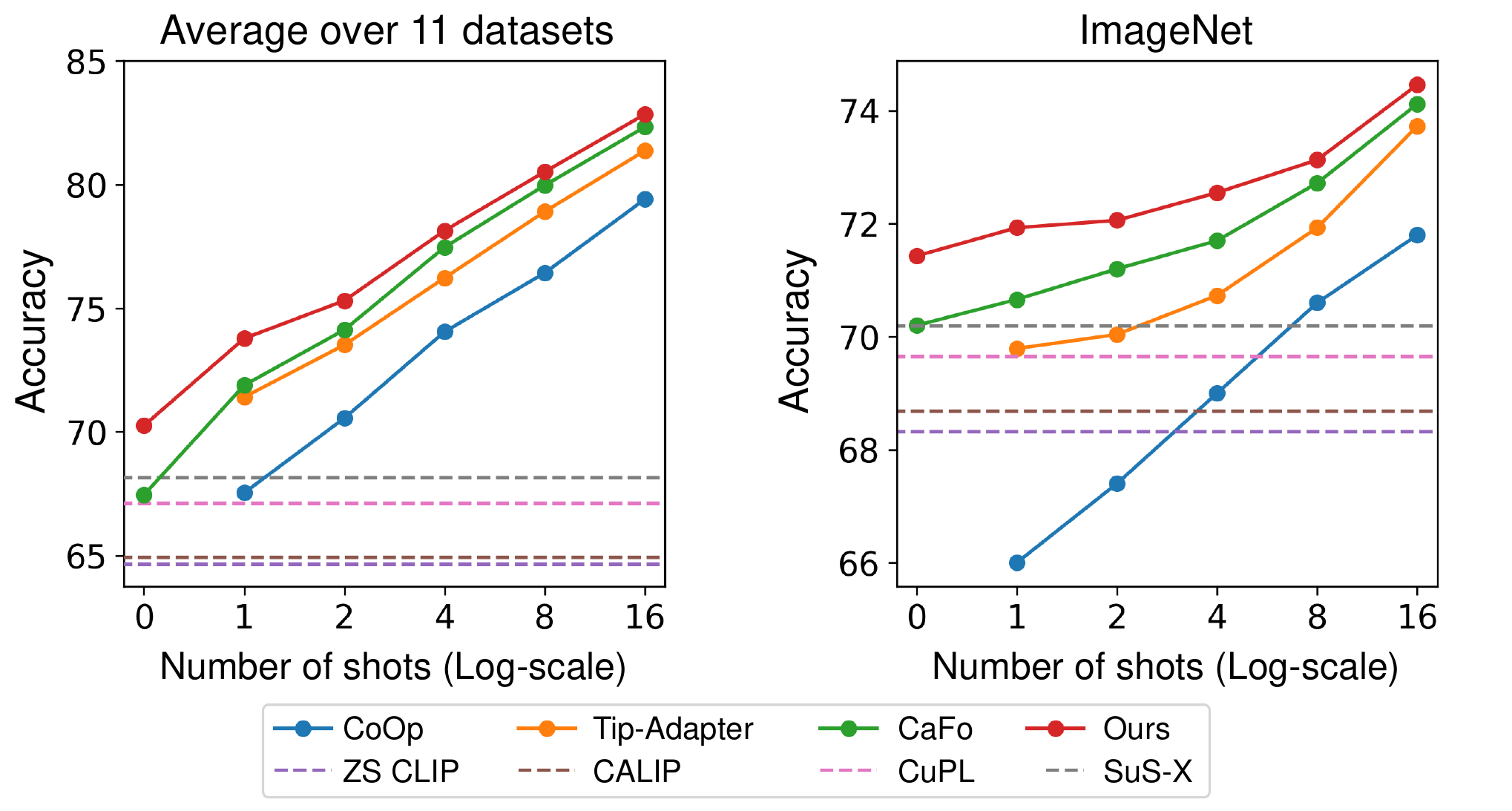}
    \caption{
    Comparison of accuracy in the few-shot classification.
    The left graph shows the average accuracy on 11 datasets, and the right graph indicates the accuracy on the ImageNet~\cite{imagenet} dataset.}
    \label{fig:exp_few_shot_results}
    \vspace{-0.2cm}
\end{figure}

\paragraph{Experimental settings and baselines}
We expand our approach to few-shot classification by upgrading the zero-shot CLIP classifier to our enhanced classifier, which was fine-tuned on the generated dataset for name-only transfer.
The generated datasets are given by Stable Diffusion 2.1~\cite{ldm} across all approaches that utilize generated datasets.
The models are trained with 1, 2, 4, 8, and 16 shots for each dataset and evaluated using the full test sets following the few-shot classification experiments~\cite{tip_adapter}.
We compare the performance with other CLIP-based adaptation approaches, which include CoOp~\cite{coop}, Tip-Adapter~\cite{tip_adapter}, and CaFo~\cite{cafo}.

\paragraph{Main results}
As shown in \Cref{fig:exp_few_shot_results}, our fine-tuned classifier consistently outperforms the baselines in the few-shot classification.
Since our proposed method aimed to utilize the generated dataset in the scenario that does not have access to real images, the performance improvements decreased with the number of real images. 
Nevertheless, our approach still demonstrates superior performance compared to few-shot baselines.
These results indicate that enhancing the CLIP image encoder using our method is also effective in data-scarce scenarios by integrating with other few-shot classification approaches.
The detailed experiments for all datasets are available in \Cref{sec:supp_full_few_shot_results}. 

\subsection{Analysis}

\begin{figure}[t]
    \centering
    \begin{minipage}{0.5\textwidth}
        \centering
        \captionof{table}{Ablation studies of the fine-tune classifier, weight-space ensemble and variance-covariance regularization on 11 datasets.}
        \vspace{0.2cm}
        \begin{tabular}{cccc}
        \toprule
        \begin{tabular}[c]{@{}c@{}}Fine-tune \\ Classifier\end{tabular} &
        \begin{tabular}[c]{@{}c@{}}Weight-space\\ Ensemble\end{tabular} &
        $\gL_{\text{VCR}}$ &
        \begin{tabular}[c]{@{}c@{}}Average\\ Accuracy\end{tabular} \\ \midrule
        \xmark & \xmark & \xmark & 67.12 \\
        \cmark & \xmark & \xmark & 53.79 \\
        \cmark & \cmark & \xmark & 69.13 \\
        \cmark & \cmark & \cmark & \textbf{69.56} \\ \bottomrule
        \end{tabular}
        \label{tab:exp_ablation}
    \end{minipage}\hfill
    \begin{minipage}{0.44\textwidth}
        \centering
         \includegraphics[width=\columnwidth]{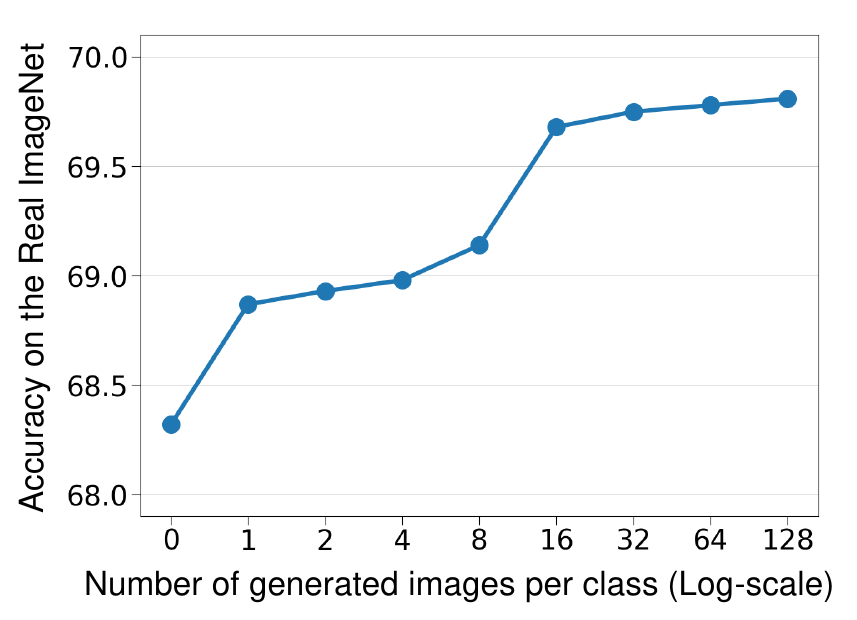}
        \caption{Real ImageNet~\cite{imagenet} accuracy according to the scale of generated datasets.}
        \label{fig:exp_num_gen_shots}
    \end{minipage}\hfill
    \vspace{-0.4cm}
\end{figure}

\paragraph{Ablation study}
The ablation study of the proposed method is conducted across all 11 datasets in the name-only transfer.
We utilize generated datasets from Stable Diffusion 2.1~\cite{ldm}, 
and we do not utilize the adapter~\cite{cafo} due to clearly showing the ablation study.
As depicted in \Cref{tab:exp_ablation}, our investigation reveals that the weight-space ensemble performs exceptionally well in overcoming the domain gap between real and generated images.
Although the fine-tuned CLIP model on the generated dataset showed significant degradation in real-world performance, the weight-space ensemble demonstrated its effectiveness by combining the zero-shot CLIP's domain-agnostic knowledge with the fine-tuned CLIP's task-specific expertise.
Furthermore, the variance-covariance regularization ($\gL_\text{VCR}$) also showed additional improvements.
The additional improvements support our analysis of feature diversity, which shows that high diversity can help improve performance in the real domain.
While the improvement from variance-covariance regularization is not substantial compared to the weight-space ensemble, both regularization methods demonstrated consistent improvements across all datasets, as shown in \Cref{sec:supp_ablation}.


\begin{table}[t]
\centering
\caption{Finding the mixing coefficient $\alpha$ for mixing between zero-shot CLIP and fine-tuned CLIP on the generated ImageNet dataset.}
\vspace{0.1cm}
\begin{tabular}{@{}M{3.0cm}M{1.0cm}M{1.0cm}M{1.0cm}M{1.0cm}M{1.0cm}M{1.0cm}M{1.0cm}@{}}
\toprule
$\alpha$ & 0.0 & 0.1 & 0.2 & 0.3 & 0.4 & $\cdots$ & 1.0 \\ \midrule
ImageNet Accuracy & 68.32 & \textbf{69.78} & 69.73 & 68.69 & 67.08 & $\cdots$ & 48.31 \\ 
{\color{blue}+ Improvement} & - & {\color{blue} \textbf{+1.46}} & {\color{blue} +1.41} & {\color{blue} +0.37} & {\color{red} -1.24} & $\cdots$ & {\color{red} -20.01} \\ 
\bottomrule
\end{tabular}%
\label{tab:exp_find_alpha}
\end{table}

\paragraph{Number of generated samples}
We conducted an analysis of the performance improvement based on the scale of the generated dataset given by Stable Diffusion 2.1~\cite{ldm} on the ImageNet~\cite{imagenet} dataset.
As illustrated in \Cref{fig:exp_num_gen_shots}, the performance on the real domain increased according to the scale of the generated datasets.
While previous name-only transfer approaches~\cite{cafo, susx} only utilized 2 generated images per class, our performance still increased by 256 images per class.
Due to our limited resources, we have ceased experimenting with more than 256 images per class.
Nevertheless, there is a potential for further scaling beyond this threshold, which could be explored using scaling laws similar to those discussed in a recent study~\cite{scaling_laws}.


\paragraph{Finding the mixing coefficient}
Determining the weight mixing coefficient is a crucial feature of the proposed method.
Unlike other hyper-parameters, such as the learning rate or the number of epochs, which require searching during training, the mixing coefficient can be effectively identified by evaluating the validation set post-training.
In our experiments, the mixing coefficient is explored in intervals of 0.1 from 0 to 1.
The accuracy on the ImageNet~\cite{imagenet} test set corresponding to the mixing coefficient $\alpha$ is presented in \Cref{tab:exp_find_alpha}.
Despite using a grid search with a relatively large interval for the mixing coefficient, it successfully integrates the zero-shot classifier and the fine-tuned classifier, demonstrating substantial performance improvements and highlighting the effectiveness of our approach.
The specific mixing coefficients for each dataset, adjusted according to the domain gap between the real and generated datasets, are available in \Cref{sec:supp_mixing_coeff}.

\section{Limitations and Discussion}
\label{sec:limitation}

\begin{figure*}[t]
    \centering
    \includegraphics[width=1.0\textwidth]{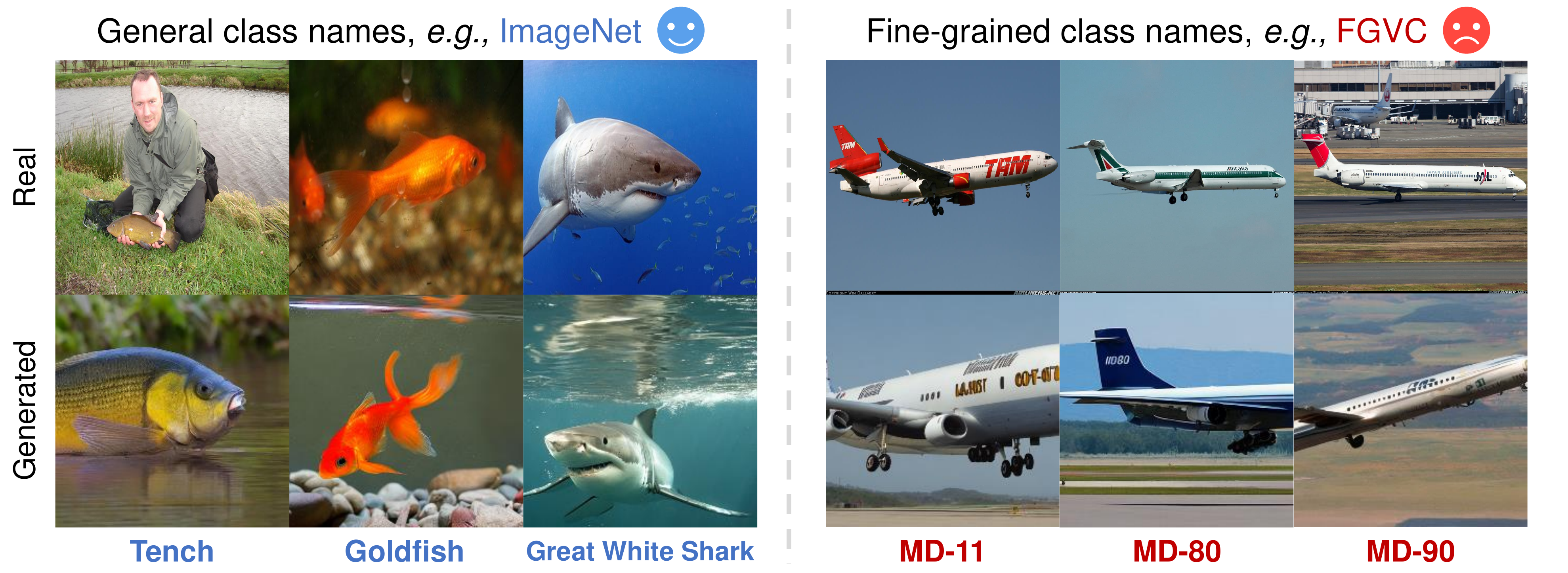}
    \caption{
    Although Stable Diffusion effectively synthesizes general images owing to its proficiency in handling general class names (\eg, ImageNet~\cite{imagenet}), it encounters challenges in generating images within specific domains from fine-grained class names (\eg, FGVC Aircraft~\cite{fgvc_aircraft}).
    }
    \label{fig:limitation_quali}
    \vspace{-0.4cm}
\end{figure*}

Although the proposed approach is the first to tackle the domain gap between real and generated images, our performance is dependent on the capability of the generative model to synthesize various subjects.
As depicted in \Cref{fig:limitation_quali}, generated images from general class names, such as ImageNet~\cite{imagenet}, exhibit high fidelity and accurately represent the target class name.
However, for narrow domains beyond the capability of Stable Diffusion, such as the FGVC~\cite{fgvc_aircraft}, the generated samples do not contain sufficient task-specific content. 
Despite the huge success of text-to-image generation, addressing the challenges of generating images for specific domains remains a crucial avenue for future research.

\section{Conclusion}
\label{sec:conclusion}

In this paper, we demonstrate the importance of addressing the domain gap between real and generated images, highlighting it as a crucial factor of performance degradation when fine-tuning exclusively with generated data.
Specifically, our in-depth analysis reveals that the primary obstacle for using generated images is not the quality of generated images (\eg, mislabeled samples), but rather the domain gap between real and generated samples.
Through our train-time and post-training regularization techniques for mitigating the domain gap, we are able to improve the image encoder itself, without using additional parameters such as adapters. 

Additionally, our findings can be seamlessly integrated into few-shot learning scenarios, also achieving state-of-the-art performance. 
While our primary experiments have focused on classification, our approach is highly adaptable to scenarios where generated images can supplement existing datasets (\eg, detection, segmentation), which we consider as promising future work.

{
\small

\bibliographystyle{plain}
\bibliography{main}

}

\clearpage


\appendix

\section*{Appendix / supplemental material}


\section{Additional details for the name-only transfer experiments}
\label{sec:supp_various_clip_results}

\paragraph{Additional implementation details}

We fine-tuned the classifier with the generated dataset for 10 epochs.
The learning rate was set to $3\times 10^{-5}$ with a cosine schedule and 500 warm-up steps.
A batch size was set to 128, and the optimizer employed was AdamW~\cite{adamw}.
The weight mixing coefficient $\alpha$ is searched in intervals of 0.1 from 0 to 1, and $\lambda_\text{Var}, \lambda_\text{Cov}$ are searched on \{0.08, 0.16, 0.32, 0.64, 1.28\}, \{0.01, 0.02, 0.04, 0.08, 0.16\}, respectively.
In our experiments, for generating 64 images for 1000 classes to construct the ImageNet dataset, it takes 2 days with 4 RTX 3090.
Then, fine-tuning the classifier with the generated ImageNet dataset takes 2 hours with a single RTX 3090.
The other datasets are in proportion to the number of the classes.
The entire training and test code will be made publicly available.

\paragraph{Licenses for existing assets}

For the datasets, we have checked that 
ImageNet (Custom (research, non-commercial))~\cite{imagenet}\footnote{\url{https://image-net.org/}}, 
Caltech101~\cite{caltech101} (CC-BY 4.0)\footnote{\url{https://data.caltech.edu/records/mzrjq-6wc02}}, 
DTD~\cite{dtd} (Custom (research-only))\footnote{\url{https://www.robots.ox.ac.uk/~vgg/data/dtd/}}, 
EuroSAT~\cite{eurosat}(MIT)\footnote{\url{https://github.com/phelber/eurosat}},
FGVCAircraft~\cite{fgvc_aircraft} (Custom (non-commercial))\footnote{\url{https://www.robots.ox.ac.uk/~vgg/data/fgvc-aircraft}},
OxfordPets~\cite{oxford_pets} (CC BY-SA 4.0)\footnote{\url{https://www.robots.ox.ac.uk/~vgg/data/pets/}},
and StanfordCars~\cite{stanford_cars} (Custom (non-commercial))\footnote{\url{https://ai.stanford.edu/~jkrause/cars/car_dataset.html}}
can be used for research purposes,
but 
Flowers102~\cite{flowers102}\footnote{\url{https://www.robots.ox.ac.uk/~vgg/data/flowers/102/index.html}},
Food101~\cite{food101}\footnote{\url{https://data.vision.ee.ethz.ch/cvl/datasets_extra/food-101/}},
SUN397~\cite{sun397}\footnote{\url{https://vision.princeton.edu/projects/2010/SUN/}},
and UCF101~\cite{ucf101}\footnote{\url{https://www.crcv.ucf.edu/data/UCF101.php}}
do not exist the license or term of use information online.
Thus, we reached out to the authors, and we are waiting for their responses.
For the pre-trained models, we have checked that
DALL-E~\cite{dalle} (Custom)\footnote{\url{https://openai.com/policies/terms-of-use/}},
DALL-E mini~\cite{dalle_mini} (Custom)\footnote{\url{https://www.craiyon.com/terms}},
Stable Diffusion~\cite{ldm} (Creative ML OpenRAIL-M)\footnote{\url{https://stablediffusion.gitbook.io/overview/stable-diffusion-overview/license}},
Stable Diffusion XL~\cite{sdxl} (Creative ML OpenRAIL-M)\footnote{\url{https://github.com/Stability-AI/generative-models/blob/main/model_licenses/LICENSE-SDXL1.0}},
and CLIP~\cite{clip} (MIT)\footnote{\url{https://github.com/openai/CLIP/blob/main/LICENSE}}
can be used for research purposes.


\paragraph{Experiments on Various CLIP Architectures}

While the CLIP ViT-B/16~\cite{clip, vit} has demonstrated high performance in the name-only transfer, it is important to note that the effectiveness of our proposed approach extends beyond this specific model architecture.
In the main paper, we consistently use the CLIP ViT-B/16 image encoder for all experiments.
In this section, we aim to illustrate that our approach is versatile across various CLIP architectures, including ResNet-50~\cite{resnet}, ResNet-101~\cite{resnet}, and ViT-B/32~\cite{vit}, as detailed in \Cref{tab:supp_various_clip}.

Our experiments reveal notable improvements with the proposed approach, particularly on ViT-based image encoders, which consistently outperform the name-only transfer baselines.
Additionally, our results indicate generally superior performance on the 11 datasets when utilizing ResNet-based image encoders.
This finding underscores the applicability of our method across diverse CLIP architectures.

\begin{table}[]
\caption{Comparison of accuracy on 11 name-only transfer benchmarks with ResNet-50, ResNet-101~\cite{resnet}, and ViT-B/32~\cite{vit} CLIP image encoder~\cite{clip}. Average indicates the average accuracy on 11 datasets. {\ul Underline} represents the highest accuracy achieved when utilizing generated datasets by DALL-E~\cite{dalle} given by CaFo~\cite{cafo}, while \textbf{Bold} indicates the highest accuracy achieved without any constraints.}
\label{tab:supp_various_clip}
\centering
\resizebox{\textwidth}{!}{%
\begin{tabular}{@{}ccccccccccccc@{}}
\toprule
\multicolumn{1}{c|}{ResNet-50} & Average & ImageNet & Caltech101 & DTD & EuroSAT & FGVC & Flowers102 & Food101 & OxfordPets & StanfordCars & SUN397 & UCF101 \\ \midrule
ZS CLIP & 56.28 & 60.20 & 85.15 & 41.84 & 28.80 & 16.95 & 61.39 & 73.22 & 81.77 & 55.74 & 58.58 & 55.46 \\
CALIP & 56.56 & 60.35 & 88.48 & 40.79 & 26.11 & 16.24 & 63.12 & 74.13 & 81.96 & 56.30 & 59.18 & 55.46 \\
CuPL & 60.82 & 61.62 & 88.28 & 50.12 & 37.48 & 20.70 & 77.64 & 64.31 & 86.43 & 57.06 & 62.25 & 63.18 \\
SuS-X (DALL-E) & 61.05 & 61.93 & 89.41 & 47.64 & 38.91 & 19.95 & {\ul 66.99} & {\ul \textbf{77.64}} & 86.35 & 56.66 & 62.71 & {\ul 63.39} \\
CaFo (DALL-E) & 61.84 & {\ul 62.48} & 90.83 & 49.47 & {\ul 42.65} & {\ul 20.58} & 65.94 & 77.54 & 86.45 & 58.66 & 62.68 & 62.94 \\
Ours (DALL-E) & {\ul 62.18} & 62.39 & {\ul \textbf{91.20}} & {\ul 50.83} & 41.64 & 20.34 & {\ul 66.99} & 77.56 & {\ul 87.22} & {\ul 60.19} & {\ul 63.06} & 62.57 \\ \midrule
CaFo (SDv2.1) & 61.30 & 62.16 & 89.74 & 49.88 & 37.49 & 20.64 & 63.91 & 77.62 & 87.00 & 59.71 & 62.77 & 63.39 \\
CaFo (SDXL) & 61.26 & 61.46 & 88.32 & 49.76 & 38.41 & \textbf{20.82} & 64.96 & 77.53 & 87.05 & 60.12 & 62.08 & 63.36 \\
Ours (SDv2.1) & \textbf{63.05} & \textbf{62.76} & 90.87 & \textbf{51.95} & \textbf{48.14} & 20.10 & 66.22 & 77.56 & 87.05 & 62.58 & \textbf{63.23} & 63.10 \\
Ours (SDXL) & 62.97 & 62.67 & 91.08 & 50.18 & 45.40 & 20.10 & \textbf{66.99} & 77.56 & \textbf{87.41} & \textbf{64.53} & 62.87 & \textbf{63.89} \\ 
\toprule
\multicolumn{1}{c|}{ResNet-101} & Average & ImageNet & Caltech101 & DTD & EuroSAT & FGVC & Flowers102 & Food101 & OxfordPets & StanfordCars & SUN397 & UCF101 \\ \midrule
ZS CLIP & 58.58 & 62.20 & 89.33 & 42.38 & 30.07 & 18.00 & 61.84 & 77.38 & 83.65 & 62.74 & 59.38 & 57.39 \\
CALIP & 58.17 & 62.33 & 90.47 & 41.67 & 24.04 & 17.16 & 62.04 & 77.67 & 83.70 & 62.73 & 59.94 & 58.13 \\
CuPL & 59.91 & 60.44 & 90.18 & 48.82 & 28.36 & 18.60 & 61.71 & 80.09 & 85.88 & 60.60 & 61.58 & 62.70 \\
SuS-X (DALL-E) & 61.87 & 62.42 & 91.89 & 48.17 & 41.51 & 18.87 & 65.49 & {\ul \textbf{80.49}} & 85.04 & 61.55 & 62.51 & 62.68 \\
CaFo (DALL-E) & 62.81 & 63.83 & 93.27 & 49.23 & 46.04 & 18.87 & 62.89 & 80.23 & 87.30 & 63.35 & 62.51 & 63.42 \\
Ours (DALL-E) & {\ul \textbf{64.09}} & {\ul 64.09} & {\ul 93.35} & {\ul \textbf{52.36}} & {\ul \textbf{48.68}} & {\ul 20.79} & {\ul \textbf{65.61}} & 80.46 & {\ul 87.98} & {\ul 65.09} & {\ul 63.10} & {\ul 63.44} \\ \midrule
CaFo (SDv2.1) & 61.90 & 63.07 & 91.72 & 49.94 & 34.40 & 19.56 & 62.28 & 80.34 & 87.76 & 66.22 & 62.77 & 62.78 \\
CaFo (SDXL) & 61.00 & 61.80 & 90.79 & 49.17 & 30.21 & 19.26 & 61.75 & 80.15 & 87.38 & 64.73 & 61.97 & 63.73 \\
Ours (SDv2.1) & 63.91 & \textbf{64.52} & \textbf{93.59} & 50.71 & 45.49 & 19.92 & 64.88 & 80.28 & \textbf{89.32} & 67.57 & \textbf{63.39} & 63.36 \\
Ours (SDXL) & \textbf{64.09} & 64.15 & 93.43 & 51.48 & 42.26 & \textbf{22.17} & 65.29 & 80.35 & \textbf{89.32} & \textbf{69.87} & 62.83 & \textbf{63.89} \\ 
\toprule
\multicolumn{1}{c|}{ViT-B/32} & Average & ImageNet & Caltech101 & DTD & EuroSAT & FGVC & Flowers102 & Food101 & OxfordPets & StanfordCars & SUN397 & UCF101 \\ \midrule
ZS CLIP & 60.61 & 64.10 & 91.76 & 43.38 & 42.17 & 18.84 & 63.78 & 78.35 & 80.49 & 60.33 & 62.19 & 61.35 \\
CALIP & 60.93 & 63.82 & 92.78 & 43.26 & 43.98 & 18.51 & 64.07 & 78.91 & 81.06 & 60.10 & 62.83 & 60.96 \\
CuPL & 64.12 & 64.92 & 92.21 & 49.76 & 47.70 & 21.78 & 67.28 & 80.84 & 89.10 & 60.83 & 65.40 & 65.56 \\
SuS-X (DALL-E) & 64.56 & 65.14 & 93.35 & 47.87 & 51.26 & 21.06 & 69.22 & 80.78 & 89.07 & 61.15 & 65.68 & 65.58 \\
CaFo (DALL-E) & 65.13 & 65.52 & {\ul 93.79} & 50.83 & {\ul 52.86} & 22.38 & 67.68 & 80.88 & 89.07 & 61.82 & 65.87 & 65.74 \\
Ours (DALL-E) & {\ul 65.75} & {\ul 65.63} & 93.63 & {\ul 52.48} & 52.41 & {\ul \textbf{22.53}} & {\ul 70.16} & {\ul \textbf{80.90}} & {\ul 89.94} & {\ul 62.77} & {\ul 66.49} & {\ul 66.32} \\ \midrule
CaFo (SDv2.1) & 64.54 & 65.18 & 93.18 & 50.24 & 47.80 & 22.11 & 66.91 & 80.81 & 89.53 & 62.67 & 65.96 & 65.50 \\
CaFo (SDXL) & 64.69 & 64.69 & 92.70 & 50.30 & 49.78 & 21.78 & 67.60 & 80.67 & 90.11 & 62.73 & 65.26 & 66.01 \\
Ours (SDv2.1) & 66.50 & \textbf{66.08} & \textbf{94.12} & \textbf{54.85} & 54.16 & 22.35 & 68.90 & 80.69 & \textbf{90.30} & 65.85 & \textbf{66.97} & 67.22 \\
Ours (SDXL) & \textbf{66.74} & 66.03 & 94.04 & 52.48 & \textbf{54.78} & 22.50 & \textbf{70.85} & 80.69 & 90.16 & \textbf{69.34} & 66.06 & \textbf{67.25} \\ \bottomrule
\end{tabular}%
}
\end{table}

\begin{table}[]
\centering
\caption{Ablation studies of the weight-space ensemble and variance-covariance regularization across all 11 datasets. It shows the general improvements of the proposed method for all datasets.}
\resizebox{\textwidth}{!}{%
\begin{tabular}{@{}ccc|cccccc@{}}
\toprule
\begin{tabular}[c]{@{}c@{}}Fine-tune \\ Classifier\end{tabular} & \begin{tabular}[c]{@{}c@{}}Weight-space\\ Ensemble\end{tabular} & $\gL_\text{VCR}$ & \multicolumn{1}{c|}{Average} & ImageNet & Caltech101 & DTD & EuroSAT & FGVC \\ \midrule
\xmark & \xmark & \xmark & \multicolumn{1}{c|}{67.12} & 69.65 & 94.28 & 54.55 & 40.70 & 27.90 \\
\cmark & \xmark & \xmark & \multicolumn{1}{c|}{53.79} & 50.18 & 77.57 & 49.00 & 50.96 & 12.75 \\
\cmark & \cmark & \xmark & \multicolumn{1}{c|}{69.13} & 70.51 & 94.28 & 57.74 & 50.96 & 27.90 \\
\cmark & \cmark & \cmark & \multicolumn{1}{c|}{\textbf{69.56}} & \textbf{70.79} & \textbf{94.93} & \textbf{57.98} & \textbf{50.99} & 27.90 \\
\midrule
\begin{tabular}[c]{@{}c@{}}Fine-tune \\ Classifier\end{tabular} & \begin{tabular}[c]{@{}c@{}}Weight-space\\ Ensemble\end{tabular} & $\gL_\text{VCR}$ & Flowers102 & Food101 & OxfordPets & StanfordCars & SUN397 & UCF101 \\
\midrule
\xmark & \xmark & \xmark & 69.47 & 86.35 & 90.54 & 66.62 & 68.00 & 70.29 \\
\cmark & \xmark & \xmark & 43.93 & 51.60 & 86.32 & 62.89 & 55.31 & 51.18 \\
\cmark & \cmark & \xmark & 70.36 & 86.35 & 91.66 & 70.77 & 69.40 & 70.45 \\
\cmark & \cmark & \cmark & \textbf{71.38} & 86.35 & \textbf{92.31} & \textbf{71.68} & \textbf{69.84} & \textbf{70.87} \\ \bottomrule
\end{tabular}%
}
\label{tab:supp_ablation_study}
\end{table}

\section{Detailed Ablation Studies Across 11 Datasets}
\label{sec:supp_ablation}

The ablation studies of the proposed method in the name-only transfer scenario are conducted across all 11 datasets as illustrated in \Cref{tab:supp_ablation_study}.
The generated datasets are synthesized by Stable Diffusion 2.1~\cite{ldm}.
Our comprehensive experiments consistently demonstrate the efficacy of the training-time and post-training regularization techniques.
The post-training regularization, weight-space ensemble, successfully overcame the performance degradation of fine-tuning the entire classifier by leveraging the zero-shot classifier.
While the weight-space ensemble does not require any additional computational cost, it shows significant and consistent improvements across all datasets.
Furthermore, the training-time regularization ($\gL_\text{VCR}$) generally shows superior performance across all datasets except two datasets by incorporating the weight-space ensemble. 

\section{Detailed Results on Few-shot Classification}
\label{sec:supp_full_few_shot_results}

\begin{figure*}
    \centering
    \includegraphics[width=\textwidth]{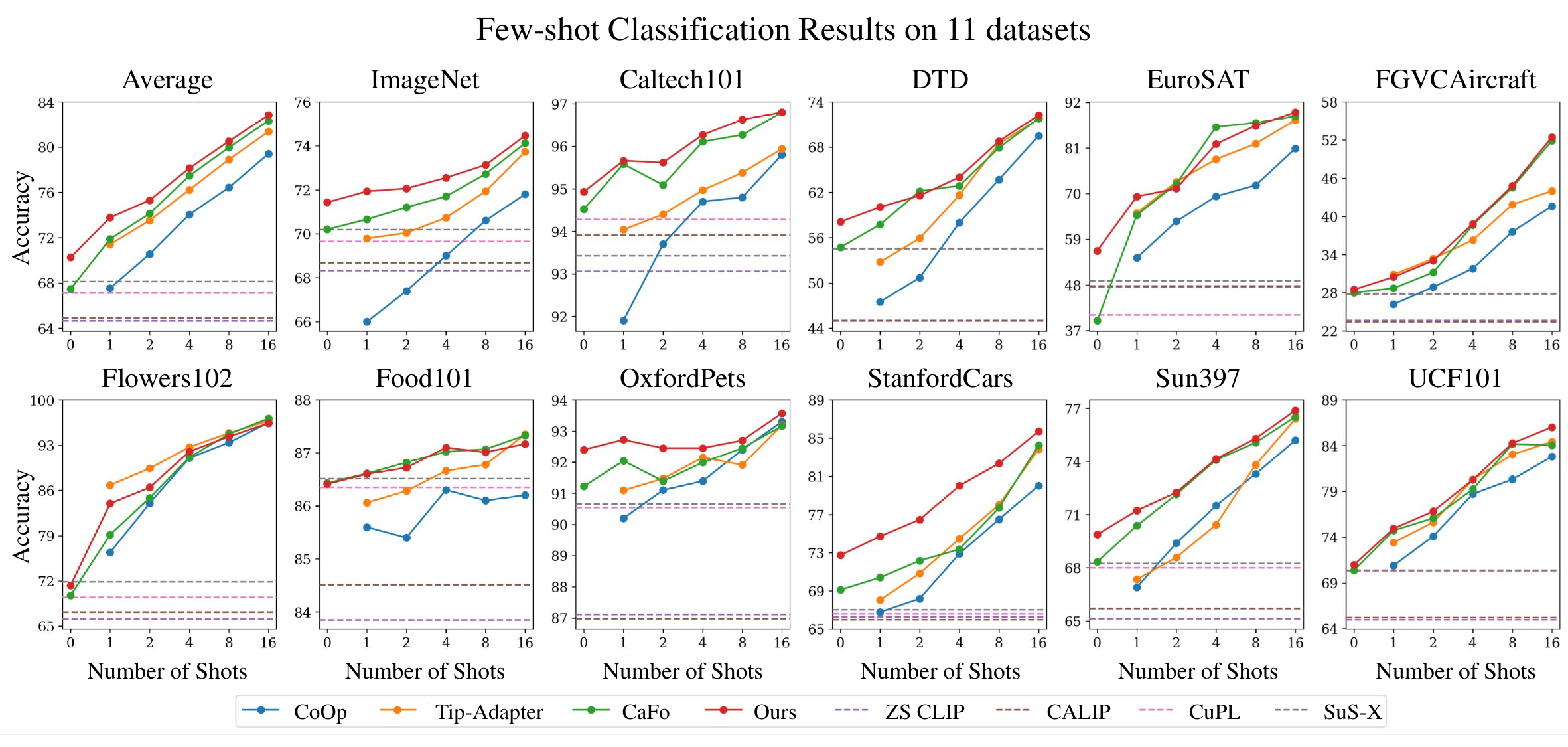}
    \caption{Comparison of accuracy in the few-shot classification setting for all 11 datasets.
    Although the enhanced CLIP image encoder has been fine-tuned on the name-only transfer scenario, the experiments show its versatility across various datasets.}
    \label{fig:supp_plot_few_shot}
\end{figure*}

In \Cref{fig:supp_plot_few_shot}, we present the few-shot classification results across all 11 datasets.
Our experiments employ CuPL~\cite{cupl} text prompts and the CaFo~\cite{cafo} adapter to achieve these outcomes.
We compare the proposed method with the adaptation of the vision-language approaches for few-shot classification: CoOp~\cite{coop}, Tip-Adapter~\cite{tip_adapter}, and CaFo~\cite{cafo}.
Additionally, the experiment incorporates name-only transfer approaches, depicted as a dotted line, following methods like CLIP~\cite{clip}, CALIP~\cite{calip}, CuPL~\cite{cupl}, and SuS-X~\cite{susx}.
The results show the effectiveness of our proposed approach in few-shot classification, even when constrained to training solely on generated datasets.

\section{Impact of Text-to-image Generation Models}
\label{sec:supp_change_text_to_image_generator}

\begin{figure}
    \centering
    \includegraphics[width=0.95\textwidth]{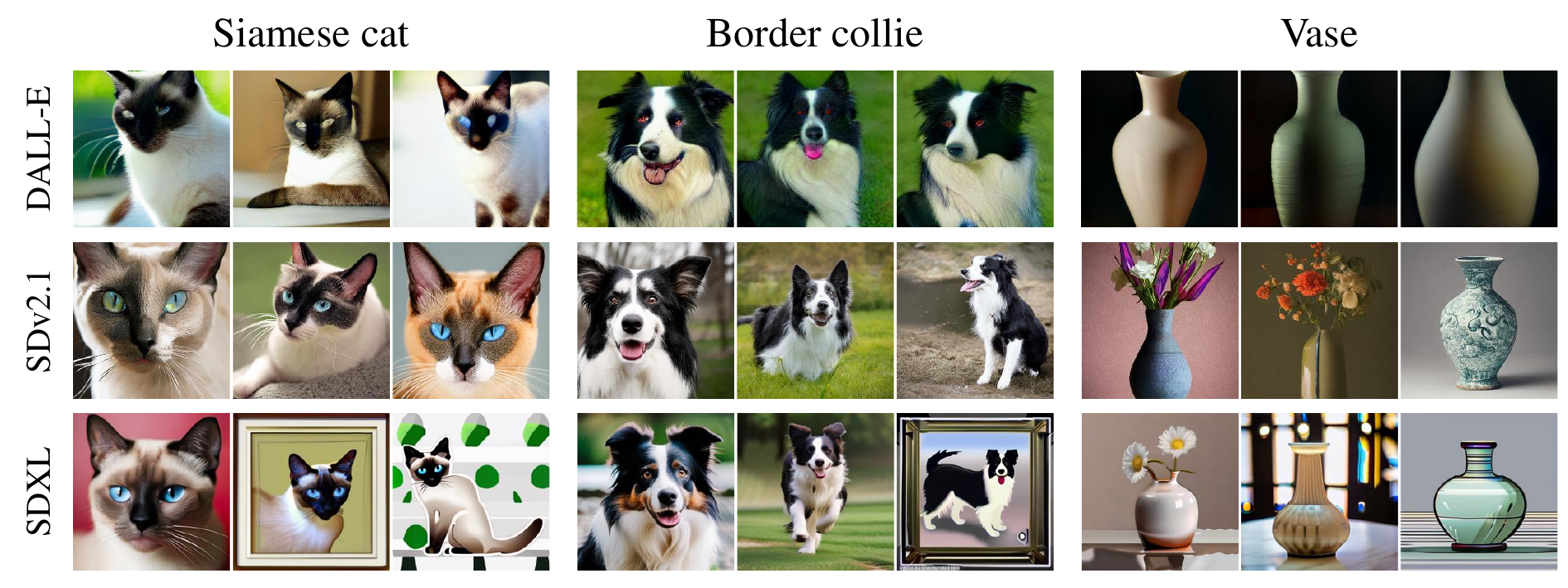}
    \caption{Qualitative results for the various text-to-image generation models. While DALL-E generates conventional images, SDv2.1 and SDXL generate in various styles.}
    \label{fig:supp_change_generation_models_quali}
\end{figure}

By scaling up the training datasets for text-to-image generation models, they can produce high-fidelity and diverse images from textual input.
Although every recent generative model effectively incorporates the textual condition, there is a noticeable difference between generative models in the diversity of the generated images.
While DALL-E~\cite{dalle} tends to generate more canonical and traditional images for each class, SDv2.1~\cite{ldm} and SDXL~\cite{sdxl} exhibit a wider range of styles from the same textual condition, as depicted in \Cref{fig:supp_change_generation_models_quali}.

\begin{figure}
    \centering
    \includegraphics[width=0.8\textwidth]{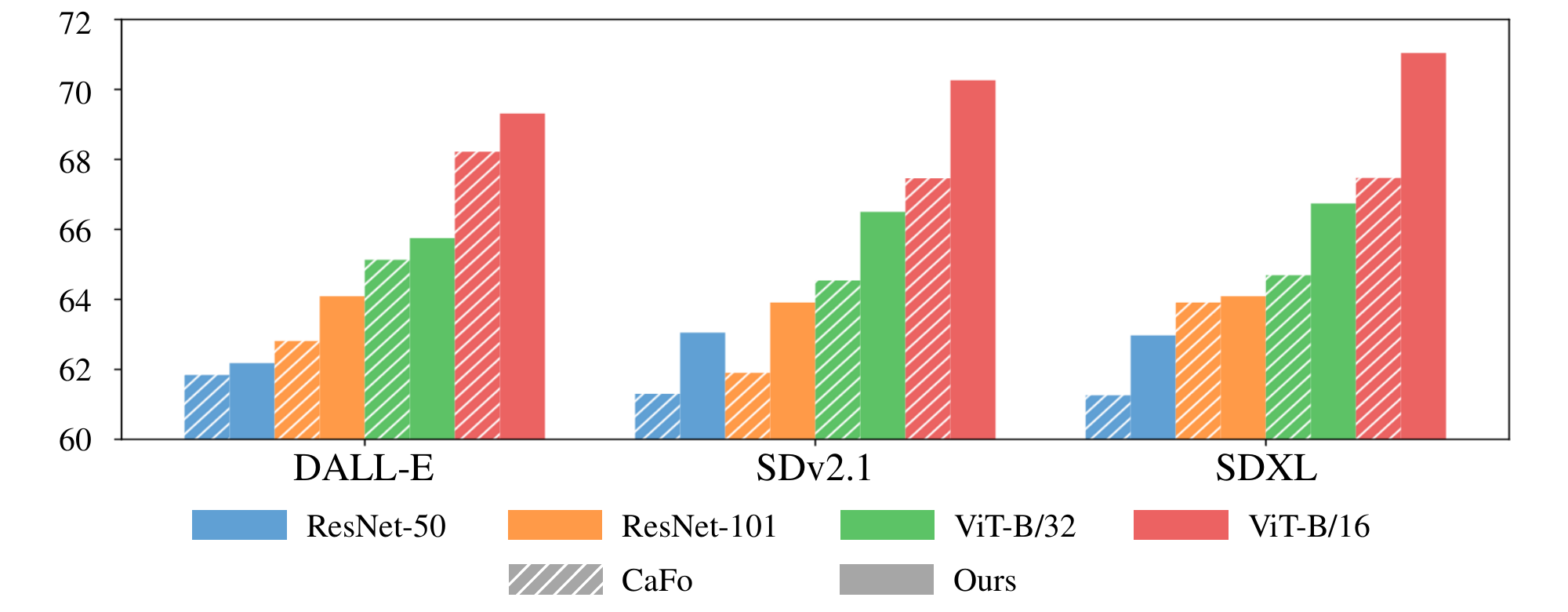}
    \caption{
    The average accuracy of the name-only transfer experiments across 11 datasets with various CLIP backbones by changing the text-to-image generation models.
    The performance gaps between CaFo and Ours increase according to the generation models.
    }
    \label{fig:supp_change_generation_models}
\end{figure}

The diversity of the generated images plays a crucial role in the name-only transfer, especially when the learnable parameters are insufficient.
Previous approaches in the name-only transfer often resorted to overcoming the domain gap between real and generated images by reducing the number of learnable parameters.
Consequently, they were unable to leverage diverse generated datasets due to their limited capacity.
In contrast, we successfully fine-tuned the entire classifier by implementing the proposed regularization techniques.
As a result, the disparity in average accuracy between CaFo~\cite{cafo} and Ours increases when employing different text-to-image generation models, as depicted in \Cref{fig:supp_change_generation_models}.

\section{Mixing Coefficient of the Weight-space Ensemble for Each Dataset}
\label{sec:supp_mixing_coeff}

The mixing coefficient $\alpha$ of the weight-space ensemble, a crucial hyper-parameter in our proposed approach, plays a pivotal role in achieving optimal results.
Through experiments across 11 datasets, we found that a fixed mixing coefficient $\alpha$ of $0.2$ consistently yields substantial performance gains from the generated datasets.
However, we recognize the need to tailor mixing coefficients individually for each dataset, considering the varying domain gaps between real and generated datasets, as illustrated in \Cref{tab:supp_mixing_coefficient}.
For example, the EuroSAT dataset~\cite{eurosat} demands a mixing coefficient $\alpha$ of $1.0$ due to the remarkable realism exhibited by the generated EuroSAT dataset, as depicted in \Cref{fig:supp_quali_1}.

As discussed in the limitations of the main paper, the weight-space ensemble, involving fine-tuned models trained on generated datasets, does not outperform the zero-shot classifier in specific domains such as FGVC Aircraft~\cite{fgvc_aircraft} and Food101~\cite{food101} due to challenges in generation quality for fine-grained classification.
Despite these limitations in fine-grained classification datasets, our experiments highlight the effectiveness of the weight-space ensemble, particularly in the name-only transfer of vision-language models, as depicted in \Cref{tab:supp_mixing_coefficient}.

\begin{table}[]
\caption{Accuracy on each dataset by changing the mixing coefficient $\alpha$. The experiments have been conducted on the final model that employs the enriched prompts~\cite{cupl} and the training-time regularization ($\gL_\text{VCR}$).}
\label{tab:supp_mixing_coefficient}
\resizebox{\textwidth}{!}{%
\begin{tabular}{@{}M{2.0cm}M{1.2cm}M{1.2cm}M{1.2cm}M{1.2cm}M{1.2cm}M{1.2cm}M{1.2cm}M{1.2cm}M{1.2cm}M{1.2cm}M{1.2cm}@{}}
\toprule
$\alpha$ & 0.0 & 0.1 & 0.2 & 0.3 & 0.4 & 0.5 & 0.6 & 0.7 & 0.8 & 0.9 & 1.0 \\ \midrule
ImageNet & 69.65 & \textbf{70.79} & 70.48 & 69.34 & 67.73 & 65.46 & 62.90 & 59.97 & 56.53 & 52.66 & 48.38 \\
Caltech101 & 94.28 & \textbf{94.93} & 94.77 & 94.44 & 94.20 & 93.75 & 92.86 & 91.52 & 90.26 & 88.15 & 85.19 \\
DTD & 54.55 & 56.44 & 57.62 & 57.92 & \textbf{57.98} & 57.86 & 56.50 & 54.96 & 53.72 & 52.07 & 50.35 \\
EuroSAT & 40.70 & 42.81 & 44.72 & 46.54 & 48.04 & 48.95 & 49.44 & 49.91 & 50.37 & 50.83 & \textbf{50.99} \\
FGVC & \textbf{27.90} & 27.00 & 26.34 & 25.11 & 23.01 & 21.66 & 20.01 & 18.39 & 17.01 & 15.66 & 14.73 \\
Flowers102 & 69.47 & 70.73 & \textbf{71.38} & 70.48 & 69.22 & 66.18 & 63.05 & 60.01 & 55.38 & 51.04 & 47.71 \\
Food101 & \textbf{86.35} & 85.91 & 84.72 & 82.62 & 79.87 & 76.34 & 72.20 & 67.47 & 62.50 & 57.29 & 51.90 \\
OxfordPets & 90.54 & 91.36 & 91.80 & 92.07 & \textbf{92.31} & 91.99 & 91.71 & 91.01 & 90.22 & 89.15 & 88.09 \\
StanfordCars & 66.62 & 69.21 & 70.64 & 71.45 & \textbf{71.68} & 71.17 & 70.63 & 69.34 & 67.98 & 65.73 & 63.31 \\
SUN397 & 68.00 & 69.36 & \textbf{69.84} & 69.45 & 68.14 & 66.49 & 64.44 & 62.30 & 59.83 & 57.17 & 54.26 \\
UCF101 & 70.29 & 70.87 & \textbf{70.87} & 70.26 & 68.94 & 67.14 & 64.37 & 61.93 & 58.68 & 55.38 & 52.50 \\ \midrule
Average & 67.12 & 68.13 & \textbf{68.47} & 68.15 & 67.38 & 66.09 & 64.37 & 62.44 & 60.23 & 57.74 & 55.22 \\ \bottomrule
\end{tabular}%
}
\end{table}

\section{Compatibility with other name-only transfer approaches}
\label{sec:supp_compatibility}

\begin{table}[t!]
\centering
\caption{Results obtained from the integration of our regularized fine-tuning method with other name-only transfer approaches by leveraging Stable Diffusion 2.1~\cite{ldm}. The blue subscripts indicate the performance improvements from the zero-shot CLIP.
}
\begin{tabular}{@{\hspace{0.25cm}}M{1.9cm}M{1.9cm}M{1.9cm}|M{2.5cm}@{\hspace{0.25cm}}}
\toprule
Regularized Fine-tuning & Enriching Prompts~\cite{cupl} & Adapter~\cite{cafo} & Average Accuracy \\ \midrule
\xmark & \xmark & \xmark & 64.66 \\
\cmark & \xmark & \xmark & 67.66 {\color{blue} \small{(+3.00)}} \\
\cmark & \cmark & \xmark & 69.55 {\color{blue} \small{(+4.89)}} \\
\cmark & \cmark & \cmark & \textbf{70.26 {\color{blue} \small{(+5.60)}}} \\ \bottomrule
\end{tabular}
\label{tab:supp_compatibility}
\end{table}

The challenge of a persistent domain gap between real and generated datasets remains a significant hurdle in the name-only transfer scenario.
As a result, previous approaches in the name-only transfer scenario have predominantly focused on enriching prompts~\cite{cupl} with large-language models~\cite{gpt3} and adapters~\cite{cafo, susx} to avoid training the numerous parameters within the CLIP image encoder.
In contrast, we successfully enhanced the CLIP image encoder by utilizing a weight-space ensemble and variance-covariance regularization on the generated image features in the name-only transfer scenario.
Departing from existing approaches, our objective of improving classification performance follows a unique path.
We showcase the performance gains of our name-only transfer method along with enriching prompts from CuPL~\cite{cupl} and adapter from CaFo~\cite{cafo} in \Cref{tab:supp_compatibility}.
In summary, our aim is to unlock the image encoder's substantial potential and showcase its compatibility with other name-only transfer approaches, ultimately enabling ours to outperform existing methods successfully.

\section{Hand-written Prompts to Synthesize Images}
\label{sec:supp_hand_written_prompts}

The prompt engineering for text-to-image generation models is one of the crucial engineering techniques for synthesizing high-quality images.
While ``a photo of a \texttt{[class name]}'' serves as a general and potent prompt for producing high-quality images, its effectiveness is notably constrained when in fine-grained classifications, \eg, EuroSAT~\cite{eurosat} and FGVC Aircraft~\cite{fgvc_aircraft}.
Therefore, we adopt the standard hand-written text prompt from CLIP followed by prior research~\cite{cafo, is_synthetic, susx}.

\begin{figure}
    \centering
    \includegraphics[width=0.9\linewidth]{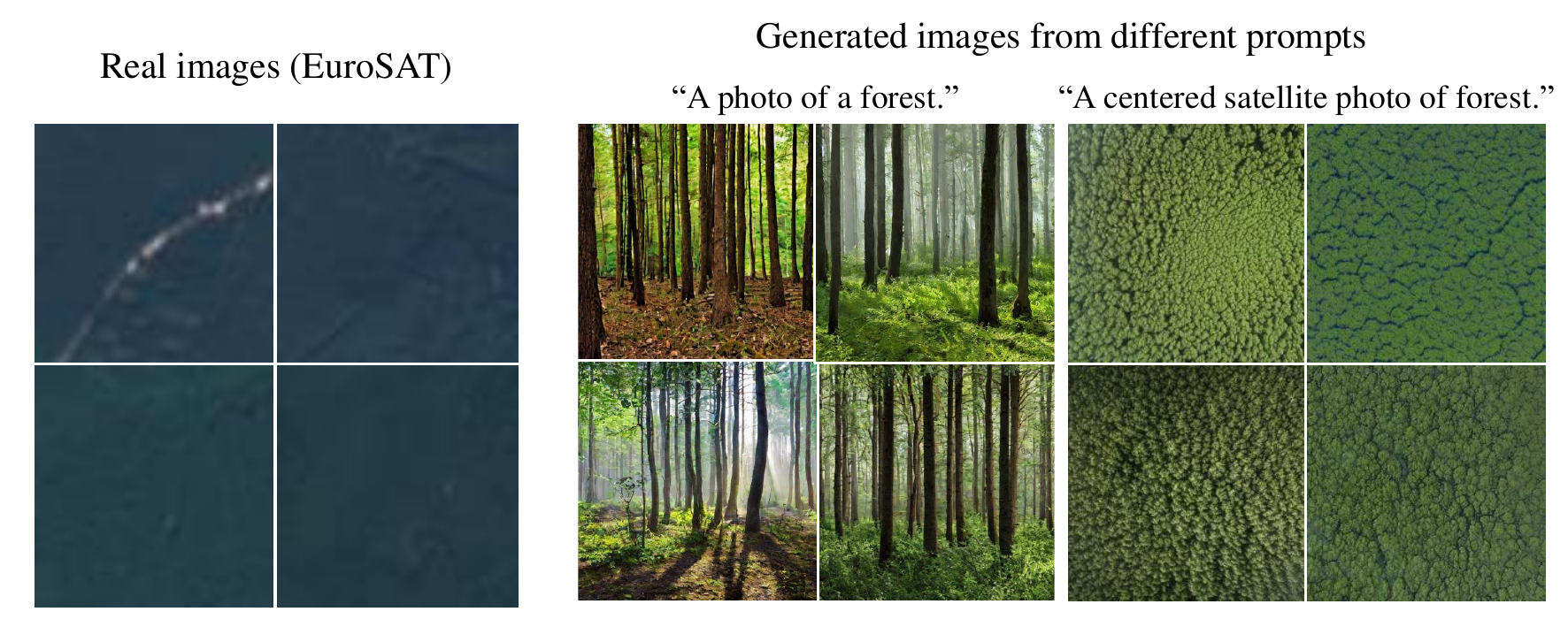}
    \caption{
    The generated images from two different prompts: ``a photo of a forest.'' and ``a centered satellite photo of a forest.''. Since the EuroSAT is the satellite domain dataset, the general prompts are limited to reducing the domain gap between the real and generated images with minimal effort.
    }
    \label{fig:supp_quali_eurosat}
\end{figure}

\begin{figure*}
    \centering
    \includegraphics[width=1.0\textwidth]{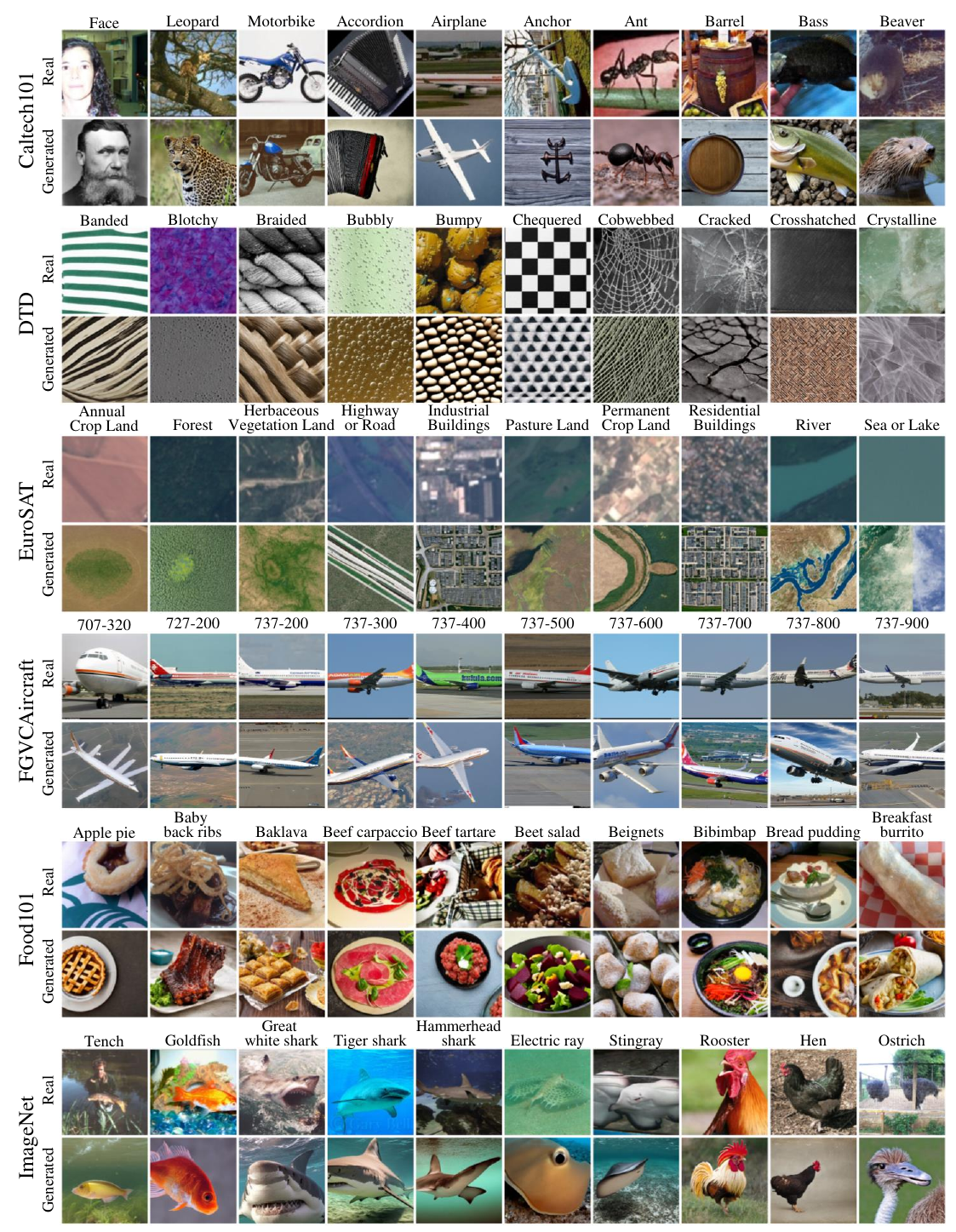}
    \caption{Qualitative results of the generated datasets. The first row contains the real images, and the second row contains the generated images for each dataset.}
    \label{fig:supp_quali_1}
\end{figure*}

\begin{figure*}
    \centering
    \includegraphics[width=1.0\textwidth]{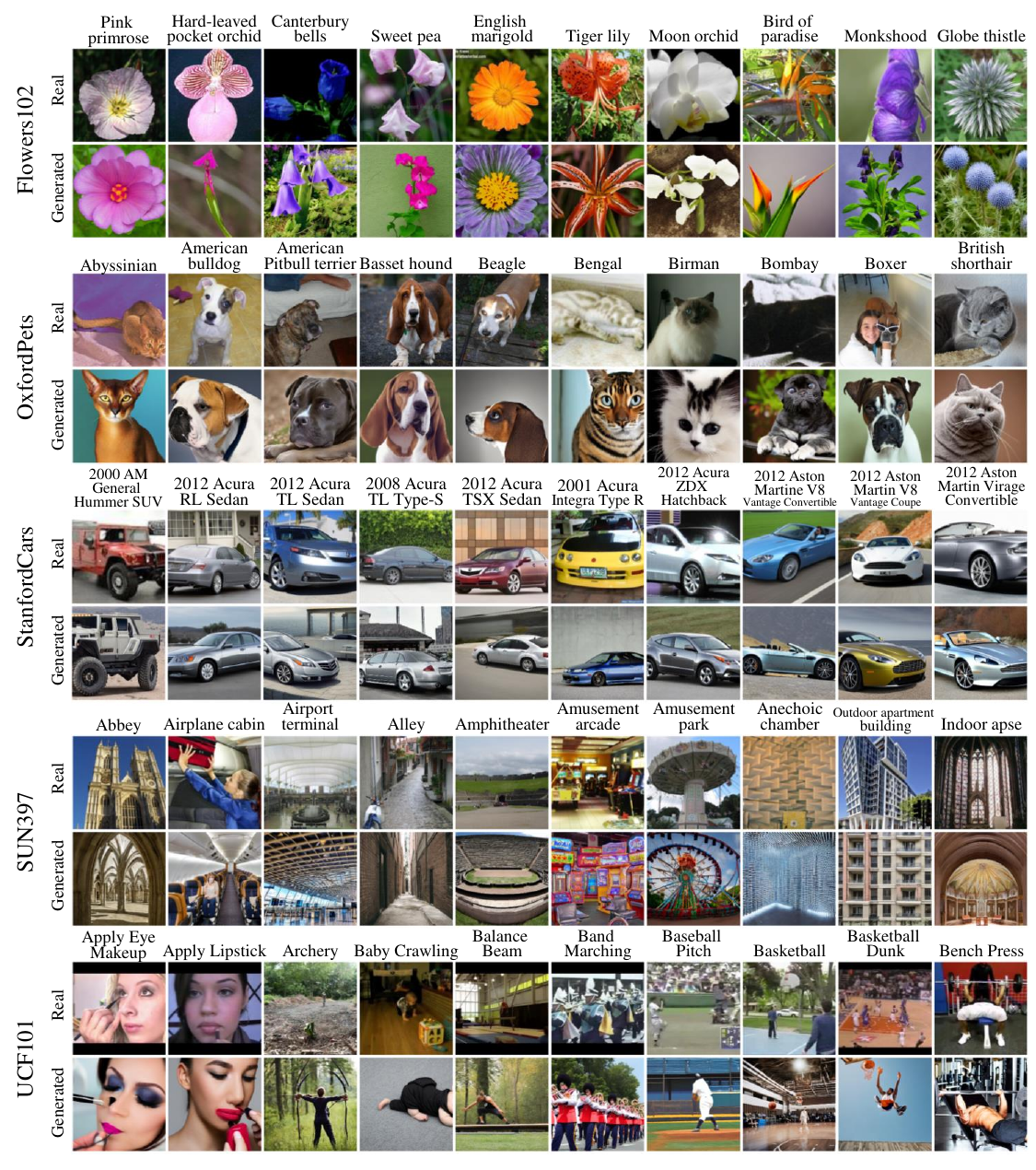}
    \caption{Qualitative results of the generated datasets. The first row contains the real images, and the second row contains the generated images for each dataset.}
    \label{fig:supp_quali_2}
\end{figure*}

We demonstrate the impact of the hand-written prompts in the synthesis of EuroSAT images.
Although the images generated by utilizing both prompts are realistic and appropriately convey the target class, the images generated from ``a photo of a forest.'' are totally different from the original EuroSAT images as shown in \Cref{fig:supp_quali_eurosat}.
Therefore, we utilize the hand-written prompts for generating datasets given by previous approaches~\cite{cafo, susx, is_synthetic}.
The specific hand-written prompts used for the evaluation of 11 datasets can be found in \Cref{tab:supp_prompts}, and the generated images for these 11 datasets are showcased in \Cref{fig:supp_quali_1,fig:supp_quali_2}.

Given the pivotal role that manual prompt engineering plays in generating datasets, exploring automatic prompt engineering techniques emerges as a crucial avenue for future research.

\begin{table}[]
\centering
\caption{Hand-written prompts used for generating images given by previous researches~\cite{cafo, is_synthetic, susx}. These hand-written text prompts are selected from the standard text prompts for zero-shot classification with CLIP.}
\begin{tabular}{@{\hspace{0.3cm}}ll@{\hspace{0.3cm}}}
\hline
Dataset & Hand-written Prompts \\ \hline
ImageNet & a photo of a \texttt{[class name]}. \\
Caltech101 & a photo of a \texttt{[class name]}. \\
DTD & a photo of a \texttt{[class name]} texture. \\
EuroSAT & a centered satellite photo of \texttt{[class name]}. \\
FGVC & a photo of a \texttt{[class name]}, a type of aircraft. \\
Food101 & a photo of \texttt{[class name]}, a type of food. \\
Flowers102 & a photo of a \texttt{[class name]}, a type of flower. \\
OxfordPets & a photo of a \texttt{[class name]}, a type of pet. \\
StanfordCars & a photo of a \texttt{[class name]}, a type of car. \\
SUN397& a photo of a \texttt{[class name]}. \\
UCF101 & a photo of a person doing \texttt{[class name]}. \\ \hline
\end{tabular}
\label{tab:supp_prompts}
\end{table}

\clearpage

\clearpage

\end{document}